\documentclass[journal]{IEEEtran}

\makeatletter
\def\ps@headings{%
	\def\@oddhead{\mbox{}\scriptsize\rightmark \hfil \thepage}%
	\def\@evenhead{\scriptsize\thepage \hfil \leftmark\mbox{}}%
	\def\@oddfoot{}%
	\def\@evenfoot{}}
\makeatother 

\makeatletter
\def\endthebibliography{%
	\def\@noitemerr{\@latex@warning{Empty `thebibliography' environment}}%
	\endlist
}
\makeatother

\usepackage{graphicx}
\usepackage{amsmath, amsfonts, epsfig, multirow, floatflt}
\usepackage{amssymb}
\usepackage{subfigure}
\usepackage{cite}
\usepackage{algorithm}
\usepackage{algorithmicx}
\usepackage{algpseudocode}
\usepackage{hyperref}
\usepackage{enumitem}
\usepackage{diagbox}
\usepackage{mathrsfs}
\usepackage{url}
\usepackage{color}

\usepackage{caption}
\usepackage{subcaption}
\usepackage{hhline}
\usepackage{array}
\usepackage{booktabs}
\usepackage{amsthm}
\usepackage{bm}
\usepackage[table,xcdraw]{xcolor}
\usepackage{verbatim}

\algblock{Input}{EndInput}
\algnotext{EndInput}
\algblock{Output}{EndOutput}
\algnotext{EndOutput}

\allowdisplaybreaks[4]
\graphicspath{{figure/}}

\begin{document}
	
 \title{Safe and Sustainable Electric Bus Charging Scheduling with Constrained Hierarchical DRL}
	
	\author{Jiaju Qi$^{1}$, Lei~Lei$^{1}$, {\it Senior Member, IEEE}, Thorsteinn Jonsson$^{2}$, Dusit Niyato$^{3}$, {\it Fellow, IEEE }
    \thanks{$^{1}$J. Qi and L. Lei are with the College of Engineering, University of Guelph, Guelph, ON N1G 2W1, Canada, {\tt\small leil@uoguelph.ca} }
    \thanks{$^{2}$T. Jonsson is with EthicalAI, Waterloo, ON N2L 0C7, Canada. }
    \thanks{$^{3}$D. Niyato is with the College of Computing and Data Science, Nanyang Technological University, 50 Nanyang Avenue, Singapore 639798. }
    }
	
	\maketitle
	
	\begin{abstract}
The integration of Electric Buses (EBs) with renewable energy sources such as photovoltaic (PV) panels is a promising approach to promote sustainable and low-carbon public transportation. However, optimizing EB charging schedules to minimize operational costs while ensuring safe operation without battery depletion remains challenging - especially under real-world conditions, where uncertainties in PV generation, dynamic electricity prices, variable travel times, and limited charging infrastructure must be accounted for. In this paper, we propose a safe Hierarchical Deep Reinforcement Learning (HDRL) framework for solving the EB Charging Scheduling Problem (EBCSP) under multi-source uncertainties. We formulate the problem as a Constrained Markov Decision Process (CMDP) with options to enable temporally abstract decision-making. We develop a novel HDRL algorithm, namely Double Actor-Critic Multi-Agent Proximal Policy Optimization Lagrangian (DAC-MAPPO-Lagrangian), which integrates Lagrangian relaxation into the Double Actor-Critic (DAC) framework. At the high level, we adopt a centralized PPO-Lagrangian algorithm to learn safe charger allocation policies. At the low level, we incorporate MAPPO-Lagrangian to learn decentralized charging power decisions under the Centralized Training and Decentralized Execution (CTDE) paradigm. Extensive experiments with real-world data demonstrate that the proposed approach outperforms existing baselines in both cost minimization and safety compliance, while maintaining fast convergence speed.


	\end{abstract}


	
	\section{Introduction}
Recent advances in sustainable transportation have emphasized the critical role of Electric Buses (EBs) in mitigating urban pollution, reducing greenhouse gas emissions, and improving public transit comfort \cite{TVT8809277,TVT9201478}. However, the electrification of bus fleets introduces significant challenges, including increased strain on local power infrastructures and rising charging costs. To address these issues, two key approaches have gained substantial attention in recent years.\par

First, power utilities have introduced dynamic pricing models featuring real-time electricity tariffs \cite{7154489} to better manage electricity demand and promote load shifting. These models enable transit operators to strategically schedule EB charging during low-price periods and, when equipped with bidirectional chargers, to participate in Vehicle-to-Grid (V2G) operations by selling electricity back to the grid during high-price periods. \par

Second, the integration of distributed renewable energy sources, such as photovoltaic (PV) panels, into bus terminal stations has emerged as a promising solution to reduce dependency on the power grid and lower operational costs. PV electricity can be directly used to charge EBs during daylight hours, thus reducing energy consumption on the grid and taking advantage of clean energy. \par

Nevertheless, the above approaches give rise to new challenges for the EB Charging Scheduling Problem (EBCSP). In general, the objective of EBCSP is to determine optimal charging decisions that minimize overall operational costs, while ensuring that all EBs maintain sufficient battery energy to complete their assigned trips on time. The integration of dynamic electricity pricing and renewable PV generation introduces two major sources of uncertainties into the smart grid environment. Electricity prices fluctuate over time due to demand management, whereas PV generation is inherently volatile due to its dependence on weather and daylight conditions. These uncertainties make it difficult to plan charging schedules that are both cost-efficient and reliable. As a result, how to optimize EB charging schedules under uncertain supply and pricing conditions, while still guaranteeing safe battery levels for uninterrupted transit operations, becomes a critical and complex problem in sustainable transportation systems. \par

In recent years, Deep Reinforcement Learning (DRL) - which combines Reinforcement Learning (RL) with Deep Neural Networks (DNNs), has demonstrated strong potential in addressing complex decision-making problems under uncertainty. Unlike traditional optimization methods that rely on explicit system modeling and accurate forecasts, DRL learns optimal policies directly through interactions with the environment, thereby eliminating the need for precise predictive models \cite{10107374}. This property makes DRL particularly suitable for the highly dynamic and uncertain EBCSP environment, where electricity prices, PV generation, and bus travel times fluctuate over time. Furthermore, DRL agents can continuously update their policies based on real-time feedback, enabling them to adapt flexibly to changes in the environment and make robust charging decisions that balance operational cost and operation safety. \par 

\subsection{Literature Review}
A considerable body of existing research has modeled the EBCSP using traditional optimization techniques, most notably as Mixed-Integer Linear Programming (MILP) problems solved with commercial solvers such as CPLEX. A comprehensive survey of these approaches can be found in \cite{zhou2024charging}. In contrast, a small but growing body of literature explores the use of DRL to address the EBCSP.\par

Among the limited DRL-based studies, Chen \emph{et al.} \cite{9014160} applied Double Q-learning (DQL) to determine the charging power for a single EB upon arrival at a terminal station. Qi \emph{et al.} \cite{11006647} presented a hierarchical algorithm to make the high-level charging target decision and low-level charging power decision. More recent works \cite{WANG2024103516,yan2024mixed,10930901,10920141} extend DRL frameworks to fleet-level coordination. Wang \emph{et al.} \cite{WANG2024103516} proposed a hybrid approach combining Clipped DQL and Soft Actor-Critic (SAC) to jointly schedule EB charging and dispatching. To reduce computational complexity, charging decisions are limited to binary variables (charge or not), sacrificing flexibility. Yan \emph{et al.} \cite{yan2024mixed} introduced a hierarchical framework integrating Twin Delayed DDPG (TD3) with an MILP, where a DRL agent assigns trips at 30-minute intervals, and an MILP module generates one-minute resolution charging plans. However, this approach assumes fixed charging power, limiting adaptability in dynamic pricing environments. \par

Distinct from the above, our previous work in \cite{10930901,10920141} addresses electricity price uncertainty and incorporates V2G capabilities to reduce charging costs. We proposed a Hierarchical DRL (HDRL) algorithm—Double Actor-Critic Multi-Agent Proximal Policy Optimization (DAC-MAPPO)—featuring a centralized high-level agent that allocates chargers at coarse time intervals in response to EB arrivals and departures. Concurrently, multiple decentralized low-level agents determine charging power at finer time resolutions (e.g., every few minutes), adapting to real-time electricity prices. \par

While existing DRL-based methods have shown promising results, they fall short in addressing two major gaps. First, none of the above studies \cite{9014160,WANG2024103516,yan2024mixed,10930901,10920141} consider renewable PV generation or address the additional uncertainty it introduces in EB charging solutions. Second, there is a lack of a systematic method to balance operational safety and cost minimization. Insufficient emphasis on safety increases the risk of battery depletion and potential in-service breakdowns, which must be strictly avoided. Conversely, overly prioritizing safety can result in conservative policies, leading to excessive charging and elevated operational costs. To ensure all EBs complete their assigned trips safely, existing studies, such as \cite{10930901}, typically include large penalty terms in the reward function to discourage battery depletion. However, tuning such penalties is non-trivial: a penalty that is too high distorts learning and leads to inefficiency; one that is too low results in frequent constraint violations. \par

Safe RL provides a principled framework to manage this trade-off by formulating the problem as a Constrained Markov Decision Process (CMDP), where safety constraints are explicitly enforced alongside the reward optimization objective \cite{8910361}. {\color{black}Leveraging primal dual methods, the constrained optimization problem can be transformed into an unconstrained dual problem, with the penalty dynamically adjusted through a learnable multiplier. This eliminates the need for manual tuning and enhances constraint satisfaction. Among existing safe RL approaches, Constrained Policy Optimization (CPO) \cite{achiam2017constrained} and Lagrangian relaxation are two representative primal–dual methods. CPO enforces constraint satisfaction during policy updates but can result in overly conservative exploration. In contrast, Lagrangian-based methods allow occasional constraint violations during exploration, yet have been shown to achieve superior final policy performance \cite{ray2019benchmarking}. This property makes them particularly attractive for training in simulated environments, where limited safety violations during learning are acceptable.}

{\color{black}In recent years, safe RL has been applied to energy-constraint vehicular systems. Existing studies have employed shielding mechanisms \cite{xu2024safe,liessner2018safe}, CMDP \cite{8910361,yang2024electric}, and model-based approaches \cite{zhu2022safe} to ensure safety under energy or power constraints, demonstrating the effectiveness of safe RL for energy management in vehicular systems. To the best of our knowledge, this work is the first to apply a hierarchical safe RL approach to the EBCSP.}

\begin{table*}
\centering
\color{black}
\caption{Contrasting this paper to the literature on EBCSP with DRL-based methods.}
\begin{tabular}{|l|c|c|c|c|c|c|} 
\hline
\textbf{Features}  &Chen \emph{et al.} \cite{9014160}&Qi \emph{et al.} \cite{11006647}&Wang \emph{et al.} \cite{WANG2024103516}&Yan \emph{et al.} \cite{yan2024mixed}&Qi \emph{et al.} \cite{10930901,10920141}& Our work  \\ 
\hline
\textbf{Fleet-level scheduling }     &&                                            & \checkmark      & \checkmark       & \checkmark &    \checkmark          \\ 
\hline
\textbf{Limited number of chargers }       &                         &                   &       & \checkmark       & \checkmark &    \checkmark          \\ 
\hline
\textbf{Adjustable charging power}      & \checkmark    &   \checkmark                                      &       &  &            \checkmark  &\checkmark      \\ 
\hline

\textbf{Uncertainty in electricity price}  &            &\checkmark                                     &    &           & \checkmark  & \checkmark         \\ 
\hline
\textbf{V2G mode}       & &\checkmark     &    & & \checkmark & \checkmark        \\ \hline
\textbf{Uncertainty in renewable energy}    && &         &    &    &    \checkmark   \\ \hline
\textbf{Safety constraint}  &  &         &     &   &  &  \checkmark         \\ 
\hline
\end{tabular}
\label{related works}
\end{table*}

\subsection{Contributions}

To address the aforementioned research gaps and leverage the strengths of DRL, this paper proposes a novel safe HDRL algorithm - DAC-MAPPO-Lagrangian - to effectively solve the EBCSP. The primary contributions of this work are summarized below, focusing on both the DRL model and the DRL algorithm. 
\begin{enumerate}
    \item \textbf{DRL Model}: We extend the DAC-based HDRL framework to model a sustainable EBCSP with integrated PV generation. This model introduces two key advancements: 
\begin{itemize}
    \item We explicitly incorporate the uncertainty of PV generation into the DRL model, along with other stochastic variables such as electricity prices and bus travel times. While prior studies have explored the integration of EBs and renewable energy, most rely on deterministic assumptions or forecast-based optimization using non-DRL techniques \cite{liu2024electric,zhang2025optimal,zaneti2022sustainable}. 

    \item To address the dual objectives of minimizing operational costs and enforcing safety constraints, we formulate the problem as a CMDP with options, and further decompose it into two augmented CMDPs - one at the high level and one at the low level - based on the DAC framework. Due to the fusion of safe RL and HRL, this formulation enables temporal abstraction and constraint-aware decision-making.
\end{itemize}

\item \textbf{DRL Algorithm}: To solve the two augmented CMDPs, we propose a novel algorithm, DAC-MAPPO-Lagrangian, which integrates the Lagrangian relaxation method into the DAC framework. Specifically:

\begin{itemize}
    \item At the high level, we employ PPO-Lagrangian to learn a centralized policy that governs charger allocation decisions. Constraint handling is explicitly incorporated into the objective through the high-level Lagrangian multiplier.
    \item At the low level, we apply MAPPO-Lagrangian to train decentralized charging power policies for individual EBs under the Centralized Training with Decentralized Execution (CTDE) paradigm. Safety constraints are enforced via the shared low-level Lagrangian multiplier, while agents learn to coordinate through global critic and constraint-adjusted advantage functions.
\end{itemize}
This algorithm ensures safe and cost-efficient charging strategies across both levels of decision-making while maintaining an efficient learning efficiency. 
\end{enumerate}

To highlight the novelty of our approach, Table~\ref{related works} provides a comparative summary of recent DRL-based methods for EBCSP. Our proposed method integrates all the listed features into a comprehensive framework.\par

The rest of the paper is organized as follows. The system model is introduced in Section II, while the CMDP model is formulated in Section III. Section IV introduces the CMDP model with options and two augmented CMDPs. The proposed algorithm is presented in Section V. Finally, our experiments are highlighted in Section VI, and conclusions are offered in Section VII.

\section{System model}

\begin{table*}[t]
\centering
\color{black}
\caption{Notation used in this paper}
\begin{tabular}[b]{p{3.0cm}<{\raggedright}p{14.0cm}<{\raggedright}}
\hline
\textbf{Notations}&\textbf{Description}\\
\hline
\specialrule{0em}{1pt}{1pt}
\multicolumn{2}{l}{\textbf{Sets}}  \\
\specialrule{0em}{1pt}{1pt}
$\mathcal{A}_t$ & The state-dependent action space at time step $t$\\
$\mathcal{I}_\omega $ & The initiation set of states for option $\omega$\\
$\mathcal{M}$, $\mathcal{M}^{\rm lay}_{t}$ &The set of EBs, the set of EBs that are currently in the layover period at time step $t$\\
$\mathcal{P}_{t}$/$\mathcal{P}_{m,t}$ & The global/local charging power action space at time step $t$\\
$\mathcal{S}$ & The state space\\
$\Omega_{t}$ & The charger allocation action space and the state-dependent option space at time step $t$\\
\specialrule{0em}{1pt}{1pt}
\multicolumn{2}{l}{\textbf{Parameters}}  \\
\specialrule{0em}{1pt}{1pt}
$B_{m,t}$ & The EB status at time step $t$ for EB $m$, $0$ for operating periods and $1$ for layover periods \\
$d$ & A small tolerance for the safety constraint violation\\
$E_{m,t}$ & The SoC level of the battery for EB $m$ at time step $t$\\
$H_t^{\rm PV}$, $H_t^{\rm price}$ & The historical PV output power and electricity prices in the period spanning from time step $t-h$ up to time step $t$\\
$K,k$ & The number of trips in a day, the index of a trip\\
$k_{m,t}$ & The trip assigned to EB $m$ at time step $t$\\
$L$, $l$ & The number of bus routes, the index of a bus route \\
$M$, $M_t$, $m$ & The number of EBs, the number of EBs in the layover period, the index of an EB\\
$N$ & The number of chargers\\
$P_t^{\rm buy}$, $P_t^{\rm sell}$ & The power purchased from and sold to the grid by the terminal station\\
$P_t^{\rm PV}$ & The power generated by PV panels\\
$\rho_t$ & The electricity price at time step $t$\\
$S_t$/$S_{m,t}$ & The global/local system state at time step $t$\\
$T$, $t$ & The number of time steps in a day, the index of a time step\\
$T^\mathrm{d}_{k_{m,t}}$ & The departure time step for the trip $k_{m,t}$\\ 
$T_m^{\rm o}$ & The duration of an operating period for EB $m$\\
$\Delta t$ & The duration of each time step\\
$\tau_{m,t}$ & The number of remaining time steps from time step $t$ to the departure time for layover periods, the number of time steps from the last departure time to time step $t$ for operating periods\\
$\lambda^{\rm safe}$ & The weight of the safety cost $c^{\mathrm{safe}}(S_t)$ \\
$\lambda_{\rm H}$, $\lambda_{\rm L}$ & The high-level and low-level Lagrange multipliers \\
$\zeta^{\rm b}$ & The weight of the battery degradation cost \\
$\zeta^{\rm s}$ & A constant value used to quantify the switching cost \\
\specialrule{0em}{1pt}{1pt}
\multicolumn{2}{l}{\textbf{Decision Variables}}  \\
\specialrule{0em}{1pt}{1pt}
$A_t$/$A_{m,t}$ & The global/local action at time step $t$\\
$p_t$/$p_{m,t}$ & The global/local charging/discharging power at time step $t$\\
$\omega_t$/$\omega_{m,t}$ & The global/local charging allocation action at time step $t$\\
\multicolumn{2}{l}{\textbf{Functions}}  \\
\specialrule{0em}{1pt}{1pt}
$\hat{A}_{t,\mathrm{H}}^{\mathrm{opr}}$, $\hat{A}_{t,\mathrm{L}}^{\mathrm{opr}}$ & The high-level and low-level operational advantage functions \\
$\hat{A}_{t,\mathrm{H}}^{\mathrm{safe}}$, $\hat{A}_{t,\mathrm{L}}^{\mathrm{safe}}$ & The high-level and low-level safety advantage functions \\
$\hat{A}_{t,\mathrm{H}}^{\mathrm{adj}}$, and $\hat{A}_{t,\mathrm{L}}^{\mathrm{adj}}$ & The high-level and low-level adjusted advantage functions \\
$c_{m,t}^{\mathrm{bat}}$/$c_{m,t}^{\mathrm{ch}}$/$c_{m,t}^{\mathrm{sw}}$ & The battery degradation/charging/switching cost for EB $m$\\
$c_{t}^{\mathrm{opr}}$ & The operational cost at time step $t$\\
$c^{\mathrm{safe}}(S_t)$ & The safety cost function\\
$C_t$ & The safety return starting from time step $t$\\
$J(\pi)$/$J^{\rm opr}(\pi)$/$J^{\rm safe}(\pi)$& The expected cumulative reward/operational reward/safety cost under policy $\pi$\\
$r(S_t, A_t)$, $r^{\rm opr}(S_t, A_t)$ & The reward function, the operational reward function\\
$V_{\pi_{\rm H}}^{\rm R}$, $V_{\pi_{\rm L}}^{\rm R}$ & The high-level and low-level reward value function \\
$V_{\pi_{\rm H}}^{\rm C}$, $V_{\pi_{\rm L}}^{\rm C}$ & The high-level and low-level cost value function \\
$R_t$ & The operational return starting from time step $t$\\
$\beta_{\omega_{t-1}}(S_t)$ & The termination condition of option $\omega_{t-1}$\\
$\varXi_{m,t}(B_{m,t},\tau_{m,t})$ & The probability that the current period is terminated at time step $t$ for EB $m$\\
$\pi$, $\pi^*$ & Policy, the optimal policy\\
$\pi_{\omega}(p_t|S_t)$ & The intra-option policy for option $\omega$\\
$\mu(\omega_t|S_t)$ & The policy over options\\
\specialrule{0.05em}{2pt}{0pt}
\end{tabular}
\label{notations}
\end{table*}

\begin{figure}[t]
\centering
\includegraphics[width=0.79\linewidth]{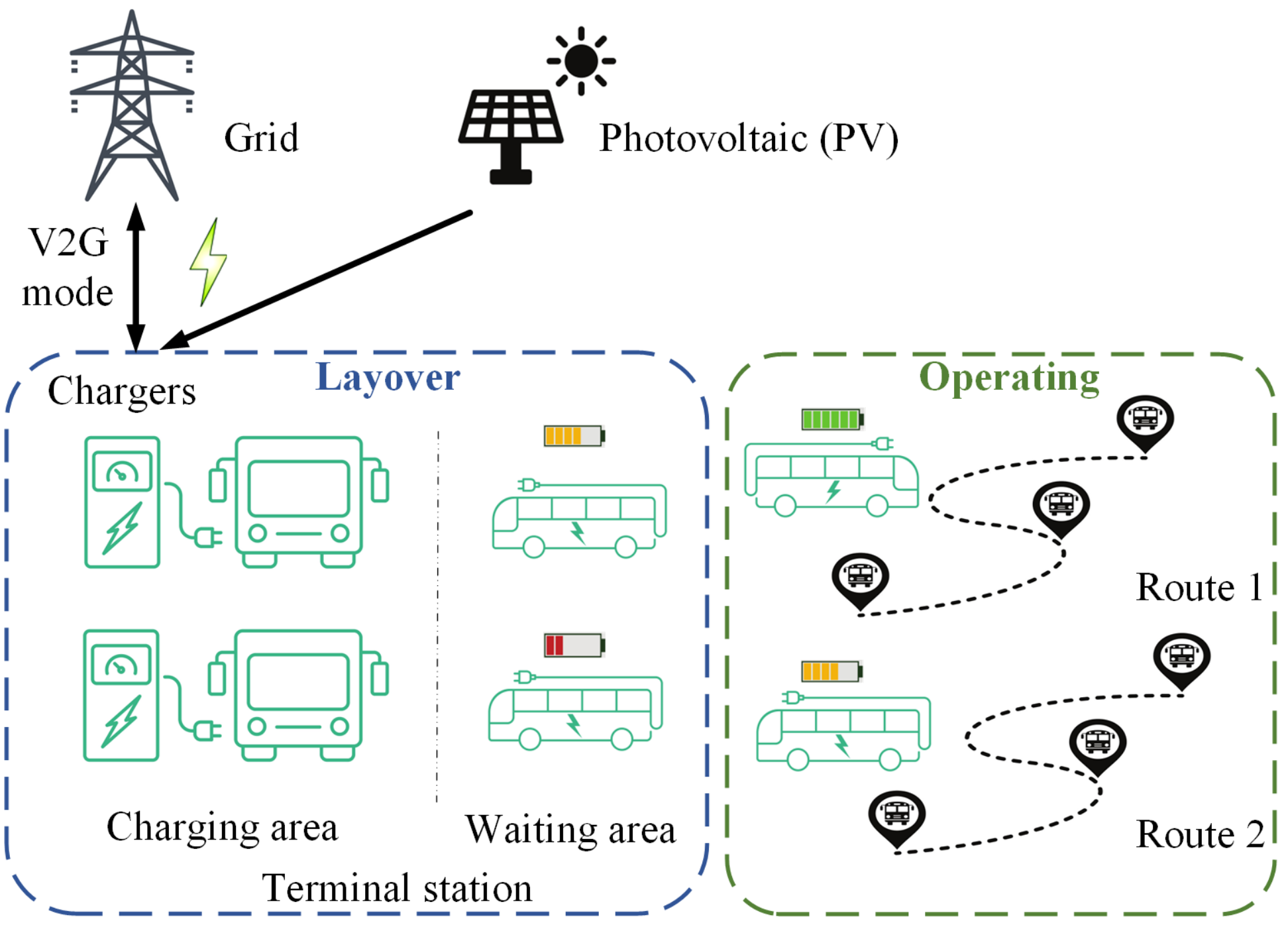}
\caption{The schematic of our system model.}
\label{terminal} 
\end{figure}

{\color{black}The notations used in this paper are summarized in Table \ref{notations}.} As shown in Fig. \ref{terminal}, we consider a sustainable terminal station equipped with solar PV panels and $N$ fast chargers, serving as a central depot for a fleet of $M$ EBs. The EBs, indexed by $m\in \{1,2,\ldots, M\}$, serve $L$ bus routes, indexed by $l\in \{1,2,\ldots, L\}$. Each route follows a fixed loop, starting and ending at the same terminal station, with a predefined schedule and multiple stops. According to the schedule, the EBs must complete $K$ trips every day, indexed by $k \in \{1,2,\ldots, K\}$. We consider that the travel time for any trip is random due to the uncertain traffic conditions.  \par

We consider a limited charging infrastructure, where the number of chargers is smaller than the number of EBs, i.e., $N < M$. To manage this constraint, the terminal station is divided into two areas, i.e., the charging area, where selected EBs connect to chargers, and the waiting area, where the remaining EBs queue for charging, following the setting in \cite{10930901,ye2022learning}. \par

The EBs operate under a V2G mode, which allows them to either charge or discharge when connected to a charger. The charging process prioritizes the utilization of renewable energy generated by PV panels. When PV generation is insufficient to meet demand, the shortfall is supplied by grid electricity. Conversely, when there is surplus renewable energy, it is fed back into the grid for sale. \footnote{{\color{black}{\color{black}This paper does not consider the deployment of energy storage system (ESS) at the charging infrastructure. While such storage can provide additional flexibility, its high capital cost may limit its use in small-scale microgrids. Therefore, we restrict our model to PV, grid, and EBs, focusing on EBs as mobile storage resources, and leave the explicit modeling of stationary ESS for future work.}}}\par

We focus on the charging scheduling problem in a single day, which is divided into $T$ equal-length time steps, indexed by $t\in\{0, \dots, T-1 \}$, with each time step having a duration of $\Delta t$.\par

\subsection{EB Operation Model}

\begin{figure}[t]
\centering
\includegraphics[width=0.8\linewidth]{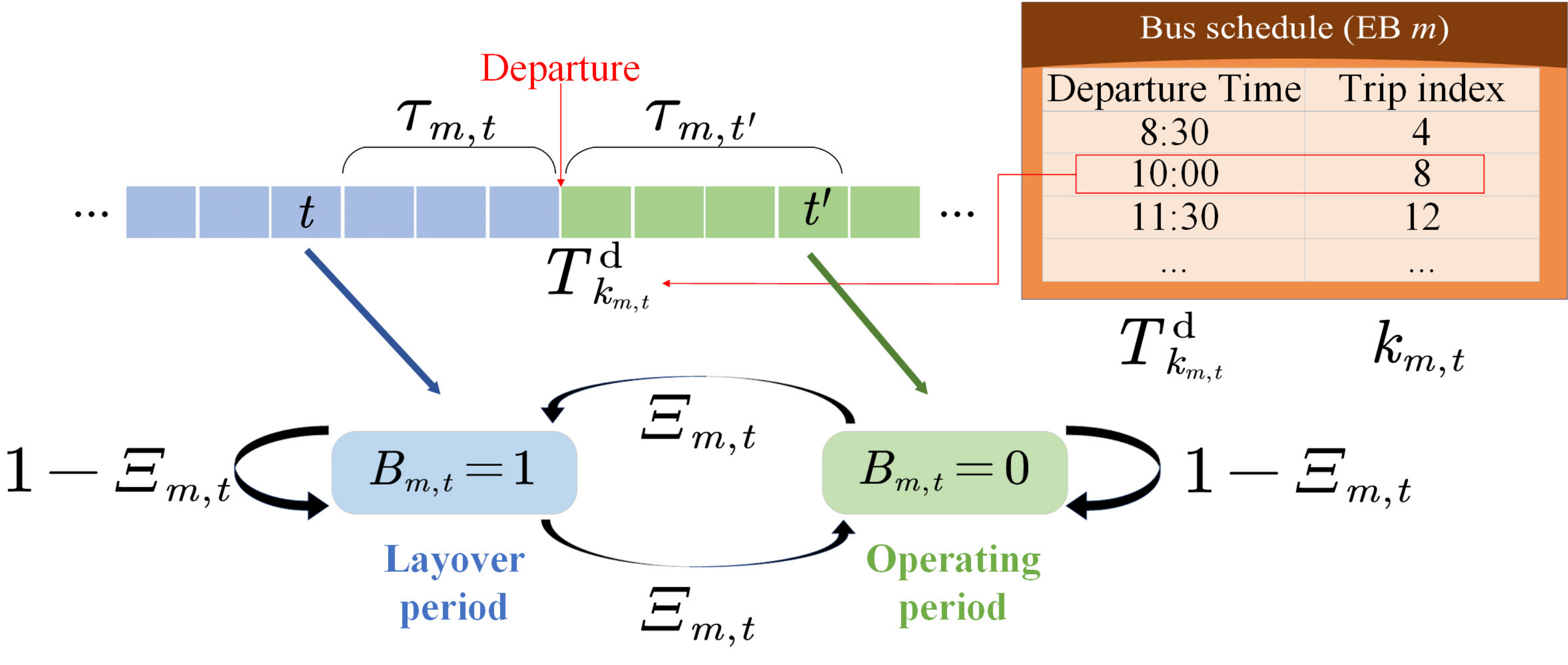}
\caption{A figure to illustrate the layover and operating periods for EB $m$ and the definitions of $B_{m,t}$, $\tau_{m,t}$, $k_{m,t}$, and $\varXi_{m,t}$.}
\label{periods} 
\end{figure}

Each EB alternates between two operational periods: a layover period, during which it remains at the terminal station, and an operating period, during which it runs along its assigned route according to a predefined schedule. Let $B_{m,t}$ be the operating status of EB $m$ at time step $t$, with $B_{m,t}=1$ for layover and $B_{m,t}=0$ for operating, as shown in Fig. \ref{periods}. \par 

Let $\varXi_{m,t}$ denote the probability that the current layover or operating period for EB $m$ is terminated at time step $t$. To derive this probability, we define a variable $\tau_{m,t}$ as 
\begin{equation}
	\label{taut}
	\\ \tau _{m,t}=\begin{cases}  T^\mathrm{d}_{k_{m,t} }-t-1,&		\mathrm{if}\,\,B_{m,t}=1 \\ \tau _{m,t-1}+1, &	\mathrm{if}\,\,B_{m,t}=0 \\\end{cases},
\end{equation}
\noindent where $k_{m,t}$ denotes the assigned trip for EB $m$ at time step $t$, and $T^\mathrm{d}_{k_{m,t} }$ represents the scheduled departure time of trip $k_{m,t}$. \par

During the layover period when $B_{m,t}=1$, $\tau_{m,t}$ represents the remaining number of time steps from time step $t$ until the departure time step $T^\mathrm{d}_{k_{m,t}}$ of trip $k_{m,t}$. In contrast, during the operating periods when $B_{m,t}=0$, $\tau_{m,t}$ denotes the number of time steps from the last departure time step of EB $m$ to the current time step $t$. \par

Note that $\varXi_{m,t}$ depends on both $B_{m,t}$ and $\tau_{m,t}$. During the operating period when $B_{m,t}=0$, the probability of EB $m$ returning to the terminal station increases as its traveling time $\tau_{m,t}$ elapses. Let $T^{\rm o}_m$ be a random variable that represents the duration of an operating period for EB $m$. The termination probability for the operating period can be expressed as
\begin{align}
	\label{gammaBt=0}
	&\varXi_{m,t}(B_{m,t}=0,\tau_{m,t})={\rm Pr}(B_{m,t+1}=1|B_{m,t}=0,\tau_{m,t}) \IEEEnonumber \\
 &=\frac{\mathrm{Pr}(T^{\rm o}_m=\tau_{m,t})}{\prod_{x=0}^{\tau_{m,t}-1}(1-\mathrm{Pr}(T^{\rm o}_m=x))}, 
\end{align}
\noindent \noindent where the numerator represents the probability of EB $m$ arriving at the terminal station at time step $t$, while the denominator represents the probability of the EB $m$ not arriving at the terminal station before time step $t$. \par

The layover period with $B_{m,t}=1$ terminates with probability $1$ when $\tau _{m,t}=0$, indicating the scheduled departure time step for EB $m$ is reached at the next time step. The termination probability for the layover period can thus be expressed as
\begin{align}
	\label{gammaBt=1}
 \varXi_{m,t}(B_{m,t}=1,\tau_{m,t})&={\rm Pr}(B_{m,t+1}=0|B_{m,t}=1,\tau_{m,t}) \IEEEnonumber \\ 
 &=\begin{cases}  1,&		\mathrm{if}\,\,\tau _{m,t}=0 	\\ 0,&		\mathrm{if}\,\,\tau _{m,t}>0 \\\end{cases}.
\end{align}

\subsection{EB Charging Model}
We use $\mathcal{M}^{\rm lay}_t$ to denote the set of EBs that are currently in the layover period at time step $t$, i.e., $\mathcal{M}^{\rm lay}_t=\{m|m\in \{1,2,\ldots, M\},B_{m,t}=1\}$. Let $M_{t}=|\mathcal{M}^{\rm lay}_t|=\sum_{m=1}^M{B_{m,t}}$ be the number of EBs in the layover period at time step $t$. \par

If $M_{t}>N$, only $N$ EBs can be charged, and the rest of the $M_{t}-N$ EBs have to enter the waiting area. Without loss of generality, we assume that the time to switch an EB from a charger to the waiting area and vice versa is negligible \cite{ye2022learning}.\par

For each EB \( m \), let \( \omega_{m,t} \in \{0,1\} \) denote its charger allocation status at time step \( t \), with \( \omega_{m,t} = 1 \) if connected to a charger, and \( 0 \) otherwise. $\omega_{m,t}$ is constrained by
\begin{equation}
\label{omega_constraint}
\sum_{m=1}^M{\omega _{m,t}\leqslant N} .
\end{equation} 

Note that only EBs in the layover period are eligible for charger allocation. When in operating period, \( \omega_{m,t} = 0 \) for any EB $m\in\mathcal{M}\backslash\mathcal{M}^{\rm lay}_t$.\par


Let $E_{m,t}$ denote the battery level of EB $m$ at time step $t$, constrained by the maximum and minimum State of Charge (SoC) level, $\eta_{\max}$ and $\eta_{\min}$, respectively. Thus, we have
\begin{equation}
    \label{Etconstraint}
    \eta_{\min}\cdot E_{\rm bat}\leqslant E_{m,t}\leqslant \eta_{\max}\cdot E_{\rm bat},
\end{equation}
\noindent where $E_{\rm bat}$ represents the battery capacity. \par

Let $p_{m,t}$ denote the charging power of EB $m$ at time step $t$. When $B_{m,t}=1$ and $\omega_{m,t}=1$, the EB can either charge energy from or discharge energy back to the electric grid in the V2G mode. When $B_{m,t}=1$ and $\omega_{i,t}=0$, the EB is in the waiting area, so $p_{m,t}=0$. When $B_{m,t}=0$, the EB is traveling and discharging, resulting in a negative value for $p_{m,t}$. Thus, the space of $p_{m,t}$, denoted as $\mathcal{P}_{m,t}$, is derived as 
\begin{equation}
	\label{ct_constraint}
	\mathcal{P}_{m,t} = 
	\begin{cases}
		\begin{aligned}
                & \left[ -p^{\rm dis}_{\max},p^{\rm ch}_{\max} \right] \cap \\ 
                & \left[ \frac{\eta_{\min} E_{\rm bat} - E_{m,t}}{\varDelta t}, \right. \\ 
			& \left. \frac{\eta_{\max} E_{\rm bat} - E_{m,t}}{\varDelta t} \right]
		\end{aligned}, 
		& \begin{aligned}
			&\text{if } B_{m,t} = 1,\\
			&\omega_{m,t} = 1
		\end{aligned} \\
		
		\{0\}, 
		& \begin{aligned}
			&\text{if } B_{m,t} = 1, \\
			&\omega_{m,t} = 0
		\end{aligned} \\
		
		\begin{aligned}
                & \left[ -p^{\rm dis}_{\max}, 0 \right] \cap \left[ \frac{ - E_{m,t}}{\varDelta t}, 0 \right]
		\end{aligned}, 
		& \text{if } B_{m,t} = 0
	\end{cases}
\end{equation}

\noindent where $p^{\rm ch}_{\max}$ and $p^{\rm dis}_{\max}$ denote the maximum absolute value of charging and discharging power, respectively. In addition, $\left[ \left( \eta_{\min}E_{\rm bat}-E_{m,t} \right) /\varDelta t,\left( \eta_{\max}E_{\rm bat}-E_{m,t} \right) /\varDelta t \right]$ represents the value range due to the limitation of EB battery capacity during the layover period. 

Finally, the dynamics of the EB battery can be modeled as
\begin{equation}
	\label{Etdynamic}
 E_{m,{t+1}}=E_{m,t}+ p_{m,t}\cdot \Delta t.
\end{equation}

\subsection{Power Balance Model}

Power balance must be satisfied at every time step, meaning that the total generated power minus the total consumed power should be equal to zero, as shown in Fig. \ref{terminal}. Thus, we have \begin{equation} P_t^{\rm buy} - P_t^{\rm sell} + P_t^{\rm PV} - \sum_{m=1}^M p_{m,t}=0, \label{power balance} \end{equation} where $P_t^{\rm buy}$ and $P_t^{\rm sell}$ represent the power purchased from and sold to the grid by the terminal station, respectively, $P_t^{\rm PV}$ denotes the power generated by PV panels, and $\sum_{m=1}^M p_{m,t}$ is the total charging power of all EBs at time step $t$. \par

The corresponding grid power transactions are determined as follows:
\begin{equation} P_t^{\rm buy} = \sum_{m=1}^M p_{m,t} - P_t^{\rm PV}, \, P_t^{\rm sell} = 0, \, \text{if } \sum_{m=1}^M p_{m,t} - P_t^{\rm PV} \geqslant 0, \label{buy} \end{equation}

\noindent and \begin{equation} P_t^{\rm buy} = 0, \, P_t^{\rm sell} = P_t^{\rm PV} - \sum_{m=1}^M p_{m,t}, \, \text{if } \sum_{m=1}^M p_{m,t} - P_t^{\rm PV} < 0. \label{sell} \end{equation}

This strategy ensures energy cost minimization while supporting the integration of renewable resources.

\subsection{Optimization Objective}

The objective function is defined as 
\begin{align}
    \label{optimization function}
    \mathrm{Minimize}\,\,\mathbb{E} \left[ \sum_{t=0}^{T-1}{\left(c_{t}^{\mathrm{opr}}+\lambda^{\rm safe}{  c_{t}^{\mathrm{safe}} }\right)} \right],
\end{align}
\noindent where $c_{t}^{\mathrm{opr}}$ denotes the operational cost at time step $t$, and $c_{t}^{\mathrm{safe}}$ is the penalty incurred when the SoC level of any EB $m\in\{1,\ldots,M\}$ falls below the minimum threshold $\eta_{\min}$ during operation at time step $t$, with
\begin{equation}
\label{Costend}
c_{t}^{\mathrm{safe}}=\sum_{m=1}^M\max \left(0,\eta_{\min}\cdot E_{\rm bat}-E_{m,t} \right).
\end{equation}

\noindent $\lambda^{\rm safe}$ is the weight that balances $c_{t}^{\mathrm{safe}}$ against $c_{t}^{\mathrm{opr}}$.\par 

$c_{t}^{\mathrm{opr}}$ can be decomposed into three components: the charging cost $c_{t}^{\mathrm{ch}}$, the battery degradation cost $c_{m,t}^{\mathrm{bat}}$, and the switching cost $c_{m,t}^{\mathrm{sw}}$, as follows:
\begin{equation}
    \label{oprcost}
    c_{t}^{\mathrm{opr}}=c_{t}^{\mathrm{ch}}+\sum_{m=1}^M{\left(  c_{m,t}^{\mathrm{bat}}+c_{m,t}^{\mathrm{sw}} \right)}.
\end{equation}

$c_{t}^{\mathrm{ch}}$ is derived as 
	\begin{equation}
		\label{chargingcost}
		c^{\mathrm{ch}}_{t}={\rho_t (P_t^{\rm buy}-\eta_{\rm sell}P_t^{\rm sell})\varDelta t},
	\end{equation}
\noindent where $\rho_t$ denotes the electricity purchase price from the grid (in \$/kWh) at time step $t$, and $\eta_{\rm sell}\in(0,1)$ is a discount factor that reflects the selling price as a proportion of the purchase price. Therefore, the effective selling price of electricity to the grid is given by $\eta_{\rm sell}\rho_t$. This pricing scheme captures the typical market condition in which the utility company purchases electricity from distributed energy resources (e.g., terminal stations) at a lower rate than that it sells, due to transmission costs and profit margins.

{\color{black}To calculate $c_{m,t}^{\mathrm{bat}}$, we adopt the linearized battery degradation model \cite{ortega2014optimal}.} Specifically, the battery degradation cost $c_{m,t}^{\mathrm{bat}}$ is positively correlated with the absolute value of the charging power $p_{m,t}$:
	\begin{equation}
		\label{batterycost}
		c_{m,t}^{\mathrm{bat}}={\zeta^\mathrm{b}\left| \frac{b_k}{100} \right|\left| \frac{p_{m,t}}{E_{\rm bat}} \right|B_{m,t}},
	\end{equation}
\noindent where $\zeta^\mathrm{b}$ is the weight of the battery degradation cost, and $b_k$ is {\color{black}the slope of the linearized battery life curve, which quantifies the degradation rate per equivalent full cycle. Theoretically, the battery capacity $E_{\mathrm{bat}}$ decreases approximately linearly with the number of charge–discharge cycles. In this study, $E_{\mathrm{bat}}$ is treated as a constant value \cite{WanZhiqiang2019MREC,PARK2022120111}, since the scheduling horizon covers only a single day and the cumulative degradation within this period is negligible.} {\color{black}Note that our proposed RL approach is compatible with more detailed battery degradation models, such as cycle-depth–dependent models \cite{7488267}.}\par

$c_{m,t}^{\mathrm{sw}}$ penalizes frequent changes in the charging status of EBs during layover periods. It is defined as
\begin{equation}
	\label{changingcost}
	c_{m,t}^{\mathrm{sw}}={\zeta^\mathrm{s}\omega_{m,t-1}\left( 1-\omega _{m,t} \right)B_{m,t}},
\end{equation}
\noindent where $\zeta^\mathrm{s}$ is a constant value used to quantify the switching cost. This cost is incurred only when an EB was connected to a charger at the previous time step ($\omega_{m,t-1}=1$) and disconnects at the current time step ($\omega_{m,t}=0$). This reflects the operational concern of avoiding unnecessary plug-in and unplugging of EBs from chargers, which may affect charging efficiency. \par

There are two decision variables: the charging power decision $p_{m,t}$, and the charger allocation decision $\omega_{m,t}$. The constraints for $p_{m,t}$ and $\omega_{m,t}$ are given by \eqref{ct_constraint} and \eqref{omega_constraint}, respectively.

\section{CMDP Model}
\subsection{MDP Formulation}

\subsubsection{State} The global state $S_t$ aggregates the local states $S_{m,t}$ for each EB $m$, i.e., $S_t=\{S_{m,t}\}_{m=1}^M$, where $S_{m,t}=\left\lbrace E_{m,t},B_{m,t},B_{m,t-1},\tau_{m,t},\omega_{m,{t-1}},H_t^{\rm PV},H_t^{\rm price},t\right\rbrace$. Let $\mathcal{S}$ denote the state space. Without loss of generality, let $\omega_{m,{-1}}=0$. Here, $H_t^{\rm PV}$ and $H_t^{\rm price}$ denote the historical PV output power and electricity prices from time step $t-h$ up to $t$, i.e., $H_t^{\rm PV}=\left( P^{\rm PV}_{t-h}, P^{\rm PV}_{t-h+1},\ldots, P^{\rm PV}_{t-1}, P^{\rm PV}_t\right)$, and $H_t^{\rm price}=\left( \rho_{t-h}, \rho_{t-h+1},\ldots, \rho_{t-1}, \rho_t\right)$,
\noindent where $h$ is the length of the time window for past data. Since $H_t^{\rm PV}$, $H_t^{\rm price}$, and $t$ are common across all $S_{m,t}$, only one instance of $H_t^{\rm PV}$, $H_t^{\rm price}$, and $t$ is included in $S_t$ after aggregation. 


\subsubsection{Action} 
At each time step $t$, the agent only determines the actions of those EBs that are currently in the layover period ($m\in\mathcal{M}^{\rm lay}_t$). Let $A_{m,t}=\{p_{m,t}, \omega_{m,t}\}$ represent the local action for EB $m$ at time step $t$. The global action $A_t$ is the aggregation of the local actions $A_{m,t}$ for all EBs in the layover period, i.e., $A_t=\{A_{m,t}\}_{m\in\mathcal{M}^{\rm lay}_t}$. Correspondingly, $A_t=(p_t,\omega_t)$ consists of two components: the charging power action, denoted by $p_t=\{p_{m,t}\}_{m\in\mathcal{M}^{\rm lay}_t}$, and the charger allocation action, denoted by $\omega_{t}=\{\omega_{m,t}\}_{m\in\mathcal{M}^{\rm lay}_t}$.  \par

Let $\mathcal{A}_t$ represent the state-dependent action space at time step $t$, where $\mathcal{A}_t=\mathcal{P}_t\times\Omega_t$. 
The charging power action space can be expressed as $\mathcal{P}_t=\Pi_{m\in \mathcal{M}^{\rm lay}_t}\mathcal{P}_{m,t}$, where $\mathcal{P}_{m,t}$ is given by \eqref{ct_constraint}. The charger allocation action space $\Omega_t$ is defined as
\begin{equation}
\label{omega_space}
\Omega_t =\left\{ \{\omega _{m,t}\}_{m\in\mathcal{M}^{\rm lay}_t}\bigg |\omega _{m,t}\in \left\{ 0,1 \right\} ,\sum_{m=1}^M{\omega _{m,t}\leqslant N} \right\}.
\end{equation} \par

The local action $A_{m,t}$ of EB $m\in\mathcal{M}\backslash\mathcal{M}^{\rm lay}_t$ during the operating period is dictated by the environment rather than determined by the agent. Specifically, the charging power action $p_{m,t}$ is a random variable whose range is specified by the third case when $B_{m,t}=0$ in \eqref{ct_constraint}. The charger allocation action $\omega_{m,t}$ is always zero. 

\subsubsection{Transition Probability}
 	The state transition probability is derived as
 	\begin{align}
 			&\mathrm{Pr}\left( S_{t+1}|S_t,A_t \right) =\mathrm{Pr}\left( H^{\rm PV}_{t+1}|H_t^{\rm PV} \right)\mathrm{Pr}\left( H^{\rm price}_{t+1}|H_t^{\rm price} \right) \IEEEnonumber \\
 			&\mathrm{Pr}\left( t+1|t \right)\prod_{i=1}^M{\bigg[ \mathrm{Pr}\left( E_{m,t+1}|E_{m,t},p_{m,t} \right)}\mathrm{Pr}\left( B_{m,t+1}|B_{m,t},\tau_{m,t} \right)   \IEEEnonumber \\
 		&\mathbf{1}_{B_{m,t}}\,\,\mathbf{1}_{\omega _{m,t}}	\,\,\mathrm{Pr}\left( \tau_{m,t+1}|\tau _{m,t},B_{m,t+1},t+1 \right)\bigg],
 		\label{pr}
 	\end{align}
 \noindent where the transition probabilities of historical PV power and electricity prices, i.e., $\mathrm{Pr}\left( H^{\rm PV}_{t+1}|H^{\rm PV}_{t} \right)$ and $\mathrm{Pr}\left( H^{\rm price}_{t+1}|H^{\rm price}_{t} \right)$ are not available, but samples of the trajectory can be obtained from real-world data. The transition probability of time steps is always $\mathrm{Pr}\left( t+1|t \right)=1$. The transition probability of the battery level for each EB, i.e., $\mathrm{Pr}\left( E_{m,t+1}|E_{m,t},p_{m,t} \right)$ can be derived from \eqref{Etdynamic}. Next, the transition probability $\mathrm{Pr}\left( B_{m,t+1}|B_{m,t},\tau_{m,t} \right)$ can be derived from the termination probability $\varXi_{m,t}(B_{m,t},\tau _{m,t})$, i.e.,
 	\begin{align}
 		\label{prtaut}
 		&\mathrm{Pr}\left( B_{m,t+1}|B_{m,t},\tau _{m,t} \right)= \\ \IEEEnonumber &\begin{cases}	1-\varXi_{m,t}(B_{m,t},\tau _{m,t}) ,&		\mathrm{if}\,\,B_{m,t+1}= B_{m,t}\\	\varXi_{m,t}(B_{m,t},\tau _{m,t}),&		\mathrm{if}\,\,B_{m,t+1}= 1-B_{m,t} \\\end{cases},
 	\end{align}
 	\noindent where $\varXi_{m,t}(B_{m,t},\tau _{m,t})$ is given by \eqref{gammaBt=1} and \eqref{gammaBt=0}. Finally, the transition probability $\mathrm{Pr}\left( \tau _{m,t+1}|\tau _{m,t},B_{m,t+1},t+1 \right)$ for $\tau_{m,t}$ is derived from \eqref{taut}. \par
 	 
\subsubsection{Reward function} 
The objective of an MDP model is to derive the optimal policy $\pi^*$ that maximizes $J(\pi)$:
\begin{equation}
	\label{objective1}
	\pi ^*=\mathrm{arg}\underset{\pi}{\max}\,J(\pi), 
\end{equation}
\noindent where $J(\pi)$ is defined as the expected cumulative reward under policy $\pi$:
\begin{equation}
     \label{return1}
    J(\pi) = \underset{\pi}{\mathbb{E}}\left[ \sum_{t=0}^{T-1}{r\left( S_t,A_t \right)} \right].
\end{equation}

Since \eqref{objective1} must align with the optimization objective in \eqref{optimization function} in Section II.D, we can derive the reward function $r\left( S_{t},A_{t} \right)$ as
\begin{equation}
	\label{rewardEB}
 r\left( S_{t},A_{t} \right) =  r^{\mathrm{opr}}\left( S_{t},A_{t} \right)-\lambda^{\rm safe} c^{\rm safe}(S_t),
 \end{equation}
\noindent where $r^{\mathrm{opr}}\left( S_{t},A_{t} \right)$ is the negative operational cost 
\begin{equation}
\label{opr-reward}
    r^{\mathrm{opr}}\left( S_{t},A_{t} \right)=-c^{\mathrm{opr}}_t,
\end{equation} 
\noindent which is a function of both state $S_{t}$ and action $A_{t}$ according to \eqref{oprcost}, \eqref{chargingcost}, \eqref{batterycost}, and \eqref{changingcost}. $c^{\rm safe}(S_t)=c^{\rm safe}_t$ is given in \eqref{Costend}, which is a function of only the state $S_{t}$. In the rest of this paper, we refer to $r^{\rm opr}\left( S_{t},A_{t} \right)$ as \textit{operational reward} and $c^{\rm safe}(S_t)$ as \textit{safety cost}. \par


Note that since $c_{t}^{\mathrm{opr}}$ depends on $c_{m,t}^{\mathrm{sw}}$ from \eqref{oprcost}, which in turn depends on $\omega_{m,{t-1}}$ according to \eqref{changingcost}, we include $\omega_{m,{t-1}}$ in the state $S_{m,t}$ defined in Section III.A.1). \par

\subsection{CMDP Formulation}
In the MDP objective function defined by \eqref{objective1}, \eqref{return1}, and \eqref{rewardEB}, we aim at maximizing the operational reward $r^{\mathrm{opr}}\left( S_{t},A_{t} \right)$ and minimizing the safety cost $c^{\rm safe}(S_t)$. A proper value of $\lambda^{\rm safe}$ plays a critical role in striking a good tradeoff between the two contradictory targets. However, it is challenging to manually find the optimal value of $\lambda^{\rm safe}$.\par  

In order to tackle this problem, we reformulate the MDP as a CMDP, treating the battery depletion penalties as safety constraints rather than components of the reward. Specifically, the CMDP reward function includes only the operational reward $r^{\mathrm{opr}}\left( S_{t},A_{t} \right)$, while $c^{\rm safe}(S_t)$ is set as the safety cost function in order to enforce the EB battery depletion constraint. \par

Thus, the objective of the CMDP is to find the optimal policy $\pi^*$ by solving the problem
\begin{equation}
\begin{aligned}
	\label{cobjective}
    & \mathbf{P}=\underset{\pi}{\max}\,J^{\rm opr}(\pi)=J^{\rm opr}(\pi^*),  \\ 
    \mathrm{s.t.} \quad &  J_{}^{\mathrm{safe}}(\pi)\leq d, 
\end{aligned}
\end{equation}
\noindent where $d$ is a small tolerance for the safety constraint violation. $J^{\mathrm{opr}}(\pi)$ denotes the expected cumulative operational reward under policy $\pi$, i.e., 
\begin{equation}
\label{opr-return}
J^{\rm opr}(\pi)=\underset{\pi}{\mathbb{E}}\left[ \sum_{t=0}^{T-1}{r^{\mathrm{opr}}\left( S_{t},A_{t} \right)} \right]=\underset{\pi}{\mathbb{E}}\left[{R}_0^{\mathrm{}}\right],
\end{equation}
\noindent where ${R}_t^{\rm }$ represents the \textit{operational return} starting from time step $t$, defined as
\begin{equation}
    \label{return-opr}
    {R}_t^{\rm }=\sum_{\iota =0}^{T-t-1}{r_{}^{\mathrm{opr}}\left( S_{t+\iota},A_{t+\iota} \right)}.
\end{equation}
Similarly, $J^{\mathrm{safe}}(\pi)$ denotes the expected cumulative safety cost under policy $\pi$, given by
\begin{equation}
\label{safe-return}
J^{\rm safe}(\pi)=\underset{\pi}{\mathbb{E}}\left[ \sum_{t=0}^{T-1}{c^{\rm safe}(S_t)} \right]=\underset{\pi}{\mathbb{E}}\left[{C}_0^{\mathrm{}}\right],
\end{equation}
where ${C}_t^{\rm }$ represents the \textit{safety return} starting from time step $t$, defined as
\begin{equation}
    \label{return-safe}
    {C}_t^{\rm }=\sum_{\iota =0}^{T-t-1}{c^{\rm safe}(S_{t+\iota})}.
\end{equation}

By constraining $J^{\mathrm{safe}}(\pi)$ within the threshold $d$, the CMDP formulation enables the agent to minimize the total operational cost while ensuring the safe operation of all EBs.

\newtheorem{theorem2}{Theorem}


\section{CMDP Model with Options}

\subsection{Options over CMDP}   
The CMDP model defined in Section III involves two types of actions that can operate at different time scales. Specifically, the charger allocation actions can persist for a variable period of time, while the charging power actions are taken per time step. In order to take advantage of the simplicities and efficiencies of temporal abstraction, we adopt the framework of options to abstract actions at two temporal levels. \par

Options can be regarded as temporally extended ``actions," which can last for multiple time steps \cite{sutton1999between}. We consider the charger allocation action $\omega_t \in \varOmega_t$ in the CMDP model as the options, where $\varOmega_t$ defined in \eqref{omega_space} becomes the state-dependent option space. An option is prescribed by the \emph{policy over options} $\mu:\mathcal{S}\times\varOmega_t\rightarrow[0,1]$, where an option $\omega_t \in \varOmega_t $ is selected according to the probability distribution $\mu(\omega_t|S_{t})$. \par

Each option $\omega\in \varOmega_t$ is associated with a triple, i.e., $\left( \mathcal{I}_{\omega} , \pi_{\omega} , \beta_{\omega} \right)$. $\mathcal{I}_{\omega}\subseteq\mathcal{S}$ is the initiation set of states, i.e., $\omega$ is available in state $S_t$ if and only if $S_t \in \mathcal{I}_{\omega}$. Considering that only the EBs in the terminal station can be charged at each time step, the initiation set $\mathcal{I}_{\omega}$ is defined as 
\begin{equation}
	\label{initI}
	\mathcal{I}_{\omega}=\left\{ S_t|B_{m,t}=1,\exists m \in \mathcal{M}\right\} .
\end{equation}

$\beta_{\omega}:\mathcal{S}^{}\rightarrow[0,1]$ denotes the termination condition for option $\omega$. At each time step $t$, the current option $\omega$, carried over from the previous time step $t-1$, terminates with probability $\beta_{\omega_{t-1}}$, or otherwise continues to the next time step with probability $1-\beta_{\omega_{t-1}}$. A new option is prescribed by $\mu$ only when the current option terminates. Note that the current option is forced to terminate when an EB departs or returns to the terminal station, as changes in $\mathcal{M}_t^{\rm lay}$ lead to a corresponding change in the action space $\mathcal{A}_t$. Therefore, we have
\begin{equation}
\label{betafunction}
\beta_{\omega_{t-1}}(S_t)=
1, \ \mathrm{if}\, \exists m \in \mathcal{M}, B_{m,t-1}+B_{m,t}=1. 
\end{equation}
\noindent Since $B_{m,t-1}$ is required to calculate \eqref{betafunction}, it is included in the state $S_{m,t}$ defined in Section III.A.1).

$\pi_{\omega}:\mathcal{S}\times\mathcal{P}_t\rightarrow[0,1]$ denotes the \emph{intra-option policy} for option $\omega$. At each time step $t$, after the option $\omega$ is determined, the charging power action $p_t$ is chosen by $\pi_{\omega}$ according to the probability distribution $\pi_{\omega}(p_t|S_{t})$.  \par

\newtheorem{remark}{Remark}
\begin{remark}[Relation between \emph{CMDP} and \emph{Options over CMDP}]
By solving the CMDP in Section III.B, a flat policy is learned to determine the charger allocation and charging power actions at each time step. In contrast, solving the options over CMDP requires learning the policy over options $\mu$, the intra-option policy $\pi_{\omega}$, and the termination condition $\beta_{\omega}$ for each option $\omega\in\varOmega_t$. The policy $\mu$ selects charger allocation options $\omega$ effective over a flexible number of time steps, as determined by $\beta_{\omega}$, which prevents prolonged charger occupation by any EB. The intra-option policy $\pi_{\omega}$ specifies the charging power at each time step under the charger allocation option $\omega$. Through the hierarchical framework, the decision-making process is decomposed into two levels. This temporal abstraction enables better exploration and faster convergence. 
\end{remark}

\subsection{Two Augmented CMDPs}
To efficiently learn $\mu$, $\pi_{\omega}$ and $\beta_{\omega}$, $\forall \omega\in\varOmega_t$, we reformulate the options over CMDP as two augmented CMDPs:  the high-level CMDP $\varPhi_\mathrm{H}$ and the low-level CMDP $\varPhi_\mathrm{L}$, based on the DAC architecture. 


\newtheorem{mydef}{Definition}
\begin{mydef}[High-Level CMDP]
	Given the intra-option policy $\pi_{\omega}$ for $\omega\in\varOmega_t$, we define the high-level MDP $\varPhi_\mathrm{H}$ as follows:
	\begin{itemize}
		\item State $S_t$. To maintain the Markov property, $S_t$ in the original MDP should be augmented by including the option of the previous time step $\omega_{t-1}$ \cite{zhang2019dac}. Since $\omega_{t-1}$ is already an element of $S_t$ by its definition, the state space of $\varPhi_\mathrm{H}$ is $\mathcal{S}_\mathrm{H}=\mathcal{S}^{}$. 
		\item Action $\omega_t$. The action in $\varPhi_\mathrm{H}$ is the option. Therefore, the action space is $\varOmega_t$.
		\item The high-level operational reward $r_{\rm H}^{\rm opr}(S_t,\omega_t)=\sum_{p\in\mathcal{P}_t}\pi_{\omega_t}(p|S_{t})r^{\rm opr}(S_t,\omega_t,p)$.
        \item The high-level safety cost $c_{\rm H}^{\rm safe}(S_t)=\sum_{p\in\mathcal{P}_t}\pi_{\omega_t}(p|S_{t})c^{\rm safe}(S_t)=c^{\rm safe}(S_t)$.  
       		\item The transition probability is defined as
 	\begin{align}
 	& \mathrm{Pr}\left( S_{t+1}|S_t,\omega_t \right) =\sum_{p\in\mathcal{P}_t}\pi_{\omega_t}(p|S_{t})\mathrm{Pr}\left( S_{t+1}|S_t,p,\omega_t \right),
 		\label{pr_high}
 	\end{align}
\noindent where $\mathrm{Pr}\left( S_{t+1}|S_t,p,\omega_t \right)$ is given in \eqref{pr}.
	\end{itemize}
\end{mydef}

The high-level policy $\pi_\mathrm{H}$ of $\varPhi _{\mathrm{H}}$ is defined as 
\begin{align}
\label{policy mu}
\pi_\mathrm{H} \left(\omega_t|S_t\right)=\left( 1-\beta_{\omega_{t-1}}(S_t) \right) \mathbb{I}_{\omega_t=\omega_{t-1}} +\beta _{\omega_{t-1}}(S_t)\mu \left(\omega_t| S_t\right) ,
\end{align}
\noindent where $\mathbb{I}$ is the indicator function. Note that $\pi_\mathrm{H}$ is a composite function of the policy over options $\mu$ and the termination condition $\beta_{\omega_{t-1}}$. Therefore, \eqref{policy mu} implies that with probability $\left( 1-\beta_{\omega_{t-1}}(S_t) \right)$ the option will remain unchanged, i.e., $\omega_t=\omega_{t-1}$, but with probability $\beta_{\omega_{t-1}}(S_t)$ it will terminate. When the option terminates, the policy $\mu \left(\omega_t| S_t\right)$ is used to generate a new option.

The high-level \textit{reward value function} $V_{\pi_\mathrm{H}}^{\rm R}(s)$ (respectively, the \textit{cost value function} $V_{\pi_\mathrm{H}}^{\rm C}(s)$) of $\pi_\mathrm{H}$ is defined as the expected cumulative reward (respectively, cost) obtained by starting from state $s$ and then following policy $\pi_\mathrm{H}$. Formally,
\begin{equation}
\label{value_reward_high}
V_{\pi_\mathrm{H}}^{\rm R}(s)=\underset{\pi_\mathrm{H}}{\mathbb{E}}\left[\sum_{t'=t}^{T-1}{r_{\rm H}^{\rm opr}(S_{t'},\omega_{t'})}\arrowvert S_{t}=s\right], \forall s\in\mathcal{S},
\end{equation}

\begin{equation}
\label{value_cost_high}
V_{\pi_\mathrm{H}}^{\rm C}(s)=\underset{\pi_\mathrm{H}}{\mathbb{E}}\left[\sum_{t'=t}^{T-1}{c_{\rm H}^{\rm safe}(S_{t'})|S_t=s}\right], \forall s\in\mathcal{S}.
\end{equation}

 The objective of the high-level CMDP is defined as
\begin{equation}
\begin{aligned}
	\label{cobjectivehigh}
    & \mathbf{P}_{\rm H}=\underset{\pi_{\rm H}}{\max}\,J_{\rm H}^{\rm opr}(\pi_{\rm H})=J_{\rm H}^{\rm opr}(\pi^*_{\rm H}),  \\
    \mathrm{s.t.} \quad &  J_{\rm H}^{\mathrm{safe}}(\pi_{\rm H})\leq d, 
\end{aligned}
\end{equation}
\noindent where  
\begin{equation}
J_{\rm H}^{\rm opr}(\pi_{\rm H})=\underset{\pi_{\rm H}}{\mathbb{E}}\left[ \sum_{t=0}^{T-1}{r_{\rm H}^{\rm opr}\left( S_{t},\omega_{t} \right)} \right],
\end{equation}
\noindent and 
\begin{equation}
J_{\rm H}^{\rm safe}(\pi_{\rm H})=\underset{\pi_{\rm H}}{\mathbb{E}}\left[ \sum_{t=0}^{T-1}{c_{\rm H}^{\rm safe}(S_t)} \right].
\end{equation}

\begin{mydef}[Low-Level CMDP]
	Given the policy over options $\mu$ and the termination condition $\beta_{\omega}$ for $\omega\in\varOmega_t$, we define the low-level MDP $\varPhi_\mathrm{L}$ as follows:
	\begin{itemize}
		\item State $(S_t,\omega_t)$. The state $S_t$ in the original CMDP is augmented by including the option of the current time step $\omega_{t}$. Thus, the state space of $\varPhi_\mathrm{L}$ is $\mathcal{S}_\mathrm{L}=\mathcal{S}^{}\times \varOmega_t$. 
		\item Action $p_t$. The action in $\varPhi_\mathrm{L}$ is the charging power action in the original CMDP. Therefore, the action space of $\varPhi_\mathrm{L}$ is $\mathcal{P}_t$.
		\item The low-level operational reward $r_{\rm L}^{\rm opr}(S_t,\omega_t,p_t)= r^{\rm opr}(S_t,A_t)$. 
                   \item The low-level safety cost is $c_{\rm L}^{\rm safe}(S_t)=c^{\rm safe}(S_t)$.
          		\item The transition probability is defined as
 	\begin{align}
 			&\mathrm{Pr}\left( S_{t+1},\omega_{t+1}|S_t,\omega_{t},p_t \right) \IEEEnonumber \\
    &=\mathrm{Pr}\left( S_{t+1}|S_t,\omega_t,p_t \right)\mathrm{Pr}\left( \omega_{t+1}|S_{t+1},\omega_t \right),
 		\label{pr_low}
 	\end{align}
    \noindent where $\mathrm{Pr}\left( S_{t+1}|S_t,\omega_t,p_t \right)\doteq\mathrm{Pr}\left( S_{t+1}|S_t,A_t \right)$ is given in \eqref{pr} and $\mathrm{Pr}\left( \omega_{t+1}|S_{t+1},\omega_t \right)\doteq\pi_\mathrm{H} \left(\omega_{t+1}|S_{t+1}\right)$ is given in \eqref{policy mu}.
    
	\end{itemize}
\end{mydef}
The low-level policy $\pi_\mathrm{L}$ of $\varPhi _{\mathrm{L}}$ is defined as 
\begin{align}
	\label{policy pi L}
	&\pi_\mathrm{L} \left( p_t|S_t,\omega_t \right) =\pi_{\omega_{t}}(p_t|S_t),
\end{align}
\noindent where $\pi_{\omega_{t}}$ is the intra-option policy of $\omega_{t}$. \par 

The low-level \textit{reward value function} $V_{\pi_\mathrm{L}}^{\rm R}(s)$ (respectively, the \textit{cost value function} $V_{\pi_\mathrm{L}}^{\rm C}(s)$) of $\pi_\mathrm{L}$ is defined as the expected cumulative reward (respectively, cost) obtained by starting from state $s$ and option $\omega$ and then following policy $\pi_\mathrm{L}$. Formally, $\forall s\in\mathcal{S}, \omega\in \varOmega_t$,
\begin{equation}
\label{value_reward_low}
V_{\pi_\mathrm{L}}^{\rm R}(s,\omega)=\underset{\pi_\mathrm{L}}{\mathbb{E}}\left[\sum_{t'=t}^{T-1}{r_{\rm L}^{\rm opr}(S_{t'}, \omega_{t'}, p_{t'})}\arrowvert (S_{t},\omega_{t})=(s,\omega)\right], 
\end{equation}

\begin{equation}
\label{value_cost_low}
V_{\pi_\mathrm{L}}^{\rm C}(s)=\underset{\pi_\mathrm{L}}{\mathbb{E}}\left[\sum_{t'=t}^{T-1}{c_{\rm L}^{\rm safe}}(S_{t'})|S_t=s\right].
\end{equation}
\newtheorem{mypropo}{Proposition}
\begin{mypropo}
 \label{Pro1}
The high-level \textit{reward value function} $V_{\pi_{\mathrm{H}}}^{\mathrm{R}}$ and \textit{cost value function} $V_{\pi_\mathrm{H}}^{\rm C}$ can be expressed in terms of the corresponding low-level \textit{reward value function} $V_{\pi_{\mathrm{L}}}^{\mathrm{R}}$ and \textit{cost value function} $V_{\pi_\mathrm{L}}^{\rm C}$ through the following equations:
\begin{equation}
    V_{\pi_{\mathrm{H}}}^{\mathrm{R}}\left( S_{t} \right) =\sum_{\omega\in \varOmega_{t}}^{}{\pi _{\mathrm{H}}\left( \omega|S_{t} \right) V_{\pi _{\mathrm{L}}}^{\mathrm{R}}\left( S_{t},\omega \right) },
    \label{value_R_highlow}
\end{equation} 
\begin{equation}
\label{value_C_highlow}
    V_{\pi_{\mathrm{H}}}^{\mathrm{C}}\left( S_{t} \right) = V_{\pi_{\mathrm{L}}}^{\mathrm{C}}\left( S_{t} \right).
\end{equation}
\end{mypropo}
\noindent The proof of Proposition \ref{Pro1} is in the Appendix A.

The objective of the low-level CMDP is defined as
\begin{equation}
\begin{aligned}
	\label{cobjectivelow}
    &\mathbf{P}_{\rm L}= \underset{\pi_{\rm L}}{\max}\,J_{\rm L}^{\rm opr}(\pi_{\rm L})=J_{\rm L}^{\rm opr}(\pi^*_{\rm L}),  \\ 
    \mathrm{s.t.} \quad &  J_{\rm L}^{\mathrm{safe}}(\pi_{\rm L})\leq d, 
\end{aligned}
\end{equation}
\noindent where  
\begin{equation}
J_{\rm L}^{\rm opr}(\pi_{\rm L})=\underset{\pi_{\rm L}}{\mathbb{E}}\left[ \sum_{t=0}^{T-1}{r^{\rm opr}_{\rm L}\left( S_{t},\omega_{t},p_t \right)} \right],
\end{equation}
\noindent and 
\begin{equation}
J_{\rm L}^{\rm safe}(\pi_{\rm L})=\underset{\pi_{\rm L}}{\mathbb{E}}\left[ \sum_{t=0}^{T-1}{c_{\rm L}^{\rm safe}(S_t)} \right].
\end{equation}

\newtheorem{remark2}[remark]{Remark}
\begin{remark2}[Relation between \emph{Options over CMDP} and \emph{two augmented CMDPs}]
When the intra-option policies $\pi_{\omega}$ are fixed, optimizing the high-level policy $\pi_{\rm H}$ implicitly involves optimizing both the policy over options $\mu$ and the termination condition $\beta_{\omega}$. Conversely, when $\mu$ and $\beta_{\omega}$ are fixed, optimizing the low-level policy $\pi_{\rm L}$ implicitly optimizes the intra-option policies $\pi_{\omega}$. Therefore, any suitable RL algorithms capable of solving CMDPs, such as DDQN \cite{vanhasselt2015deepreinforcementlearningdouble}, DDPG \cite{lillicrap2019continuouscontroldeepreinforcement}, and PPO \cite{chen2018adaptive}, can be independently and iteratively applied to address the high-level and low-level augmented CMDPs, which leads to the solution of the Options over CMDP. 
\end{remark2}

\section{DAC-MAPPO-Lagrangian Algorithm}

\begin{algorithm}[t]
	\caption{The DAC-MAPPO-Lagrangian algorithm}
	\label{alg-DAC-MAPPO}
	\begin{algorithmic}[1]  
		\State Randomly initialize the high-level actor network ${\mu}(\omega_t|S_t;\theta)$ with $\theta_0$, the low-level actor network ${\pi}_{\rm L}(p_{m,t}|S_{m,t},\omega_t,m;\vartheta)$ with $\vartheta_0$, the reward critic network ${V}_{\rm R}(S_t,\omega_t;{\phi_{\rm R}})$ with $\phi_{\rm R 0}$, the cost critic network ${V}_{\rm C}(S_t,\omega_t;{\phi_{\rm C}})$ with $\phi_{\rm C0}$, and the termination condition network $\beta_{\omega_{t-1}}(S_t;\varphi)$ with $\varphi_0$. Initialize Lagrange multipliers $\lambda_{\mathrm{H}0},\lambda_{\mathrm{L}0}$. 
		
        \For{each iteration}
        \State Initialize the buffer $\mathcal{B}$
        \For{episode $e = 1$ to $E$}
		\State Initialize the start state $S_t\gets S_0$ and the termination condition $\beta \gets 1$
		\For{time step $t = 0$ to $T-1$}
		\State Calculate the high-level policy $\pi_{\rm H}(\omega_t|S_t;\theta,\varphi)$ with $\beta$ and ${\mu}(\omega_t|S_t;\theta)$ according to \eqref{policy mu}
		\State Sample option $\omega_{t}\sim \pi_{\mathrm{H}}(\omega_t|S_t;\theta,\varphi)$
		\For{each EB $m\in \{1,\dots,M\}$ }
		\If{$\omega_{m,t}=1$}
		\State Sample local action $p_{m,t} \sim{\pi}_{\rm L}(p_{m,t}|S_{m,t},\omega_t,m;\vartheta)$ 
		\Else
		\State Observe $p_{m,t}$ from environment
		\EndIf
  \EndFor
  \State Execute low-level actions $p_{t}=\{p_{m,t}\}_{m=1}^{M}$ 
  \State Observe operational reward $r^{\rm opr}(S_{t},A_t)$, safety cost $c^{\rm safe}(S_t)$, and next state $S_{t+1}$ 
  \State Store trajectories into $\mathcal{B}$
  \State Set $\beta=\beta_{\omega_{t}}(S_{t+1};\varphi)$
		\EndFor
 \State Compute ${R}^{\rm }_t$ and ${C}^{\rm }_t$ according to \eqref{return-opr} and \eqref{return-safe}
  \State Compute $\hat{A}_{t,\mathrm{H}}^{\mathrm{opr}}$, $\hat{A}_{t,\mathrm{L}}^{\mathrm{opr}}$, $\hat{A}_{t,\mathrm{H}}^{\mathrm{safe}}$, and $\hat{A}_{t,\mathrm{L}}^{\mathrm{safe}}$ using GAE  
  \State Compute adjusted advantage functions $\hat{A}_{t,\mathrm{H}}^{\mathrm{adj}}$ and $\hat{A}_{t,\mathrm{L}}^{\mathrm{adj}}$ according to \eqref{adj_advantageH} and \eqref{adj_advantageL}
  \EndFor
  \State Update $\theta$ and $\varphi$ by maximizing the objective $L_{\mathrm{H}}^{\mathrm{ppo}}\left( \theta,\varphi \right)$ in \eqref{ppo-loss}, via mini-batch SGA
  \State Update $\vartheta$ by maximizing the objective $L_{\mathrm{L}}^{\mathrm{mappo}}\left( \vartheta \right)$ in \eqref{loss-mappo}, via mini-batch SGA  
  \State Update $\phi_{\rm R}$ and $\phi_{\rm C}$ by minimizing mean-squared errors in \eqref{loss-critic-R} and \eqref{loss-critic-C}, via mini-batch SGD 
  \State Update $\lambda_{\rm H}$ and $\lambda_{\rm L}$ according to \eqref{update-lambdaH} and \eqref{update-lambdaL}
	\EndFor	
	\end{algorithmic}  
\end{algorithm} 

\begin{figure}[t]
\centering
\includegraphics[width=0.95\linewidth]{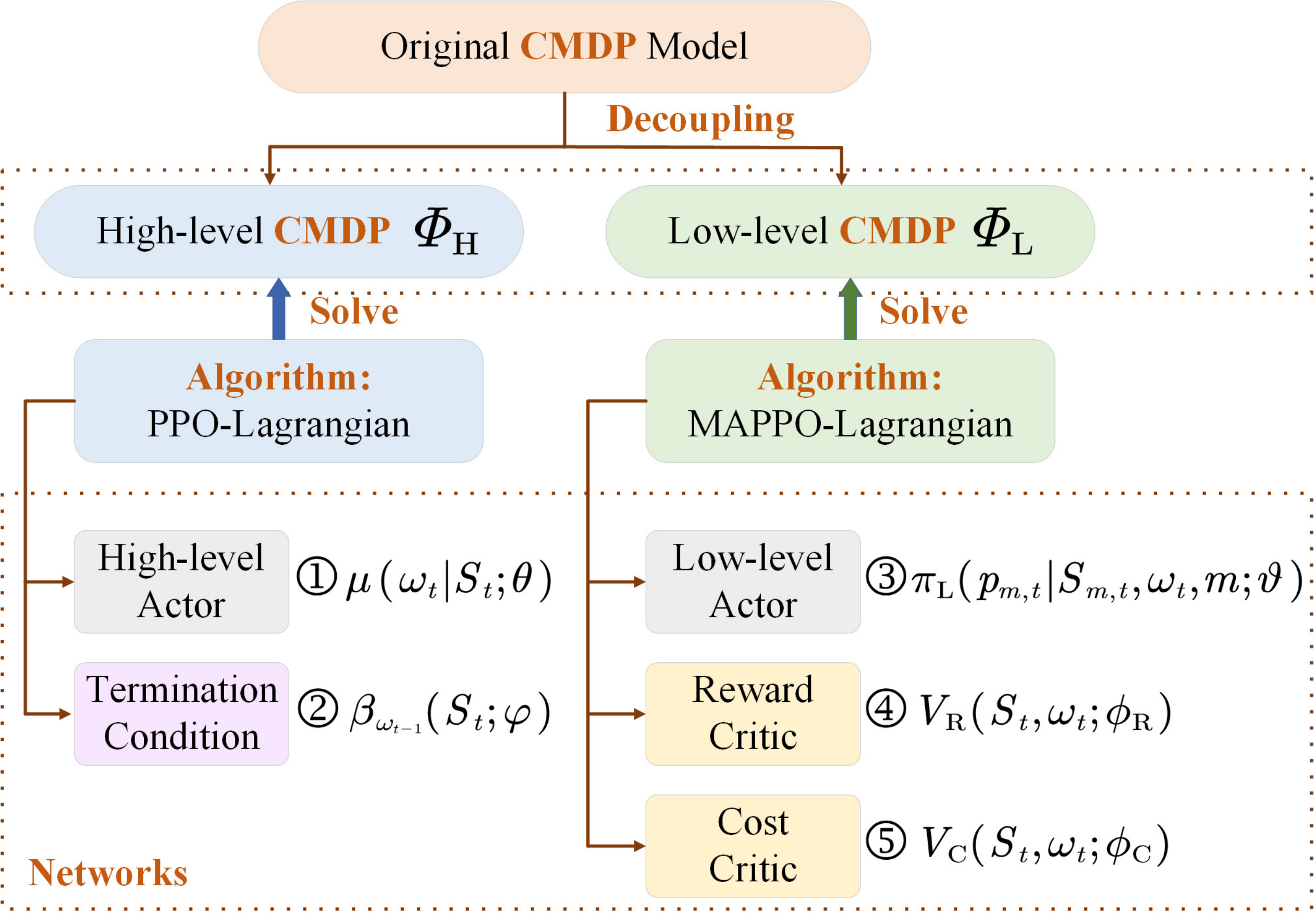}
\caption{Framework of our proposed algorithm. }
\label{algorithm-framework} 
\end{figure}

Building upon the DAC architecture, we propose a new safe HDRL algorithm, i.e., DAC-MAPPO-Lagrangian, to solve the two hierarchical augmented CMDPs presented in Section IV.B. The framework of our presented algorithm is shown in Fig. \ref{algorithm-framework}.

\subsection{DAC-CTDE architecture}

At the high level, we employ a single centralized agent that observes the global state $S_t$ and learns the high level policy $\pi_{\rm H}(\omega_t|S_t)$. As $\pi_{\rm H}$ is a compound policy that integrates both $\mu(\omega_t|S_t)$ and $\beta_{\omega_{t-1}}(S_t)$ according to \eqref{policy mu}, the high-level agent uses two distinct networks: the high-level actor network ${\mu}(\omega_t|S_t;\theta)$ with parameter $\theta$ and the termination condition network $\beta_{\omega_{t-1}}(S_t;\varphi)$ with parameter $\varphi$, to approximate $\pi_{\mathrm{H}}(\omega_t|S_t;\theta,\varphi)$. 

At the low level, we employ $M$ decentralized agents, treating each EB as an agent that observes its local augmented state $(S_{m,t},\omega_t)$. This approach not only enhances scalability as the number of EBs increases but also achieves faster convergence by reducing the dimensionality of the state space \cite{10930901}. Since the termination condition learned at the high level influences the charging duration for the allocated EBs at the low level, the low-level charging power decisions of each EB are not entirely independent, which leads to inevitable interactions among the EBs. \par

To effectively manage the multi-agent cooperation between the low-level agents, the CTDE framework is adopted. The low-level local policy $\pi_{m,{\rm L}}\left( p_{m,t}|S_{m,t},\omega_t \right)$ for an EB $m$ is derived using a decentralized actor network. Parameter sharing technique is applied, where all low-level agents share the same actor network ${\pi}_{\rm L}(p_{m,t}|S_{m,t},\omega_t,m;\vartheta)$ with parameter $\vartheta$ to expedite the training. \par

Additionally, a pair of centralized critic networks is used to approximate the reward and cost value functions: a reward critic network ${V}_{\rm R}(S_t,\omega_t;\phi_{\rm R})$ with parameter $\phi_{\rm R}$ approximates the low-level reward value function $V_{\pi_{\mathrm{L}}}^{\mathrm{R}}(S_t,\omega_t)$, and a cost critic network ${V}_{\rm C}(S_t,\omega_t;\phi_{\rm C})$ with parameter $\phi_{\rm C}$ approximates the low-level cost value function $V_{\pi_{\mathrm{L}}}^{\mathrm{C}}(S_t,\omega_t)$. Since the high-level reward and cost value functions can be expressed in terms of the corresponding low-level value functions according to Proposition 1, a single pair of reward and cost critic networks is sufficient to serve both the high-level and low-level policies.\par

For computational efficiency, we optimize both the high-level and low-level policies using the same experience samples collected from interactions with the environment. \par
\subsection{Learning centralized high-level policy using PPO-Lagrangian} 
The centralized high-level agent is responsible for solving the high-level augmented CMDP to learn optimal charger allocation decisions while explicitly enforcing safety constraints. To effectively solve this CMDP, we utilize the Lagrangian relaxation method {\color{black}along with the PPO algorithm \cite{chen2018adaptive}}. \par

Formally, we define the Lagrangian function:
\begin{equation}
\mathcal{L}_{\mathrm{H}}\left(\pi_{\mathrm{H}}, \lambda_{\mathrm{H}}\right)=J_{\mathrm{H}}^{\mathrm{opr}}\left(\pi_{\mathrm{H}}\right)-\lambda_{\mathrm{H}}\cdot\left[J_{\mathrm{H}}^{\mathrm{safe}}\left(\pi_{\mathrm{H}}\right)-d\right],
\end{equation}
where $\lambda_{\mathrm{H}}$ is the Lagrange multiplier. {\color{black}$\lambda_{\mathrm{H}}$ can be viewed as a dynamic penalty coefficient that adjusts adaptively during training — it increases when safety constraints are violated to discourage unsafe actions, and decreases when the constraints are satisfied to avoid excessive penalization.}\par 

The CMDP objective can be converted to the following unconstrained optimization objective:
\begin{equation}
\label{minmax_high}
\mathbf{D}_{{\rm H}}=\underset{\lambda _{\mathrm{H}}\geq 0}{\min}\underset{\pi _{\mathrm{H}}}{\max}\mathcal{L} _{\mathrm{H}}\left( \pi _{\mathrm{H}},\lambda _{\mathrm{H}} \right)=\mathcal{L} _{\mathrm{H}}\left( \pi^* _{\mathrm{H}},\lambda^* _{\mathrm{H}}\right).
\end{equation}

To solve the minimax problem \eqref{minmax_high}, we employ the iterative primal dual method, where the high-level policy $\pi_{\mathrm{H}}(\omega_t|S_t)$ and the Lagrange multiplier $\lambda_{\mathrm{H}}$ are updated alternately at each training iteration:

\begin{itemize}
    \item We first fix $\lambda_{\rm H}$ to update $\pi_{\mathrm{H}}(\omega_t|S_t;\theta,\varphi)$. We aim at optimizing the PPO clipped surrogate objective function $L_{\mathrm{H}}^{\mathrm{ppo}}\left( \theta,\varphi \right)$, defined as 
\begin{align}
\label{ppo-loss}
L_{\mathrm{H}}^{\mathrm{ppo}}\left( \theta,\varphi \right) =\frac{1}{ET}\sum_{e=1}^E\sum_{t=0}^{T-1}\min \left\{ \mathrm{ratio}_{\mathrm{H}}({\theta,\varphi})\cdot \hat{A}_{t,\mathrm{H}}^{\mathrm{adj}} ,\right.\\ \IEEEnonumber
\left.\mathrm{clip}\left( \mathrm{ratio}_{\mathrm{H}}({\theta,\varphi}),1- \varepsilon, 1+\varepsilon \right) \cdot \hat{A}_{t,\mathrm{H}}^{\mathrm{adj}}\right\} ,
\end{align}
\noindent where $E$ is the number of episodes executed in each iteration. Additionally,
\begin{equation}
    \label{ratioH}
\mathrm{ratio}_{\mathrm{H}}({\theta,\varphi})=\frac{\pi _{\mathrm{H}}\left( \omega_t|S_t ;{\theta,\varphi}\right)}{\pi _{\mathrm{H}}\left( \omega_t|S_t;{\theta_\mathrm{old},\varphi_\mathrm{old}} \right)},
\end{equation}
and $\hat{A}_{t,H}^{\mathrm{adj}}$ denotes the high-level adjusted advantage function calculated by:
\begin{equation}
\label{adj_advantageH}
    \hat{A}_{t,\mathrm{H}}^{\mathrm{adj}} =\hat{A}_{t,\mathrm{H}}^{\mathrm{opr}} -\lambda_{\rm H} \cdot \hat{A}_{t,\mathrm{H}}^{\mathrm{safe}}. 
\end{equation}
In \eqref{adj_advantageH}, $\hat{A}_{t,\mathrm{H}}^{\mathrm{opr}}$ and $\hat{A}_{t,\mathrm{H}}^{\mathrm{safe}}$ represent the high-level operational and safety advantage functions, respectively. These advantage functions can be calculated based on the high-level reward and cost value functions, i.e., $V_{\pi_{\mathrm{H}}}^{\mathrm{R}}(S_t)$ and $V_{\pi_{\mathrm{H}}}^{\mathrm{C}}(S_t)$, using Generalized Advantage Estimator (GAE) \cite{schulman2015high}. \par
    
As discussed in Section IV.A, $\pi_{\mathrm{H}}(\omega_t|S_t)$ is a compound policy approximated simultaneously by two networks ${\mu}(\omega_t|S_t;\theta)$ and $\beta_{\omega_{t-1}}(S_t;\varphi)$ with parameters $\theta$ and $\varphi$, respectively. The parameters $\theta$ and $\varphi$ are updated to maximize $L_{\mathrm{H}}^{\mathrm{ppo}}\left( \theta,\varphi \right)$ via mini-batch stochastic gradient ascent (SGA). \par

\item We then fix the high-level policy $\pi_{\mathrm{H}}(\omega_t|S_t;\theta,\varphi)$ and update the Lagrangian multiplier $\lambda_{\mathrm{H}}$ according to 
\begin{align}
\label{update-lambdaH}
    \lambda_{\mathrm{H}}\gets \mathrm{ReLU}\left( \lambda _{\mathrm{H}}-\alpha _{\lambda}\cdot \nabla _{\lambda _{\mathrm{H}}}\left( \mathcal{L}_{\mathrm{H}}\left( \pi_{\mathrm{H}},\lambda_{\mathrm{H}} \right) \right) \right) \\ \IEEEnonumber
    =\mathrm{ReLU}\left( \lambda_{\mathrm{H}}+\alpha _{\lambda}\cdot \left( J_{\mathrm{H}}^{\mathrm{safe}}\left( \pi _{\mathrm{H}} \right) -d \right) \right) ,
\end{align}
\noindent where $\alpha_{\lambda}$ is the learning rate for updating $\lambda_{\mathrm{H}}$, and $\mathrm{ReLU}(\cdot)$ denotes the projection onto the non-negative dual space, ensuring that $\lambda_{\mathrm{H}} \geq 0$.
\end{itemize}

\subsection{Learning decentralized low-level policy using MAPPO-Lagrangian}
The decentralized low-level agents are responsible for solving the low-level augmented CMDP to learn optimal charging power decisions while explicitly enforcing safety constraints. To effectively solve this CMDP, we also utilize the Lagrangian relaxation methods {\color{black}along with the MAPPO algorithm \cite{yu2022surprising} }. The Lagrangian function is defined as
\begin{equation}
\mathcal{L}_{\mathrm{L}}\left(\pi_{\mathrm{L}}, \lambda_{\mathrm{L}}\right)=J_{\mathrm{L}}^{\mathrm{opr}}\left(\pi_{\mathrm{L}}\right)-\lambda_{\mathrm{L}}\cdot\left[J_{\mathrm{L}}^{\mathrm{safe}}\left(\pi_{\mathrm{L}}\right)-d\right],
\end{equation}
\noindent {\color{black}where $\lambda_{\mathrm{L}}$ is the Lagrange multiplier, serving a similar purpose as $\lambda_{\mathrm{H}}$.}\par 
The minimax problem is derived as
\begin{equation}
\label{minmax_low}
    \mathbf{D}_{\rm L}=\underset{\lambda _{\mathrm{L}}\geq 0}{\min}\underset{\pi _{\mathrm{L}}}{\max}\mathcal{L}_{\mathrm{L}}\left( \pi_{\mathrm{L}},\lambda_{\mathrm{L}} \right)=\mathcal{L}_{\mathrm{L}}\left( \pi^*_{\mathrm{L}},\lambda^*_{\mathrm{L}} \right).
\end{equation}

To solve the problem in \eqref{minmax_low}, we also employ the iterative primal dual method, where each agent $m$ updates its own decentralized low-level policy $\pi_{m,{\rm L}}\left( p_{m,t}|S_{m,t},\omega_t \right)$, while the Lagrange multiplier $\lambda_{\mathrm{L}}$, the reward critic ${V}_{\rm R}(S_t,\omega_t;{\phi_{\rm R}})$, and cost critic ${V}_{\rm C}(S_t,\omega_t;{\phi_{\rm C}})$ are updated in a centralized manner. \par 

Specifically, at each training iteration:

\begin{itemize}
    \item We first fix $\lambda_{\rm L}$ to update $\pi_{\mathrm{L}}(p_{m,t}|S_{m,t},\omega_{t},m;\vartheta)$. We aim at optimizing the PPO clipped surrogate objective function $L_{\mathrm{L}}^{\mathrm{mappo}}\left( \vartheta \right)$, defined as

\begin{align}
\label{loss-mappo}
L_{\mathrm{L}}^{\mathrm{mappo}}\left( \vartheta \right) =\frac{1}{ETM}\sum_{e=1}^E\sum_{t=0}^{T-1}\sum_{m=1}^M{\min \left\{ \mathrm{ratio}_{m,\mathrm{L}}(\vartheta)\cdot  \right.}\\ \IEEEnonumber
\left.\hat{A}_{t,\mathrm{L}}^{\mathrm{adj}},\mathrm{clip}\left( \mathrm{ratio}_{m,\mathrm{L}}(\vartheta),1- \varepsilon,1+\varepsilon \right) \cdot \hat{A}_{t,\mathrm{L}}^{\mathrm{adj}}\right\} ,
\end{align}
where
\begin{equation}
 \mathrm{ratio}_{m,\mathrm{L}}(\vartheta)=\frac{\pi _{\mathrm{L}}\left( p_{m,t}|S_{m,t},\omega_{t},m;\vartheta \right)}{\pi _{\mathrm{L}}\left( p_{m,t}|S_{m,t},\omega_{t},m ;{\vartheta_\mathrm{old}}\right)},
    \label{ratioL}
\end{equation}

\begin{equation}
\label{adj_advantageL}
    \hat{A}_{t,\mathrm{L}}^{\mathrm{adj}} =\hat{A}_{t,\mathrm{L}}^{\mathrm{opr}} -\lambda_{\rm L} \cdot \hat{A}_{t,\mathrm{L}}^{\mathrm{safe}}.
\end{equation}

The parameter $\vartheta$ is updated to maximize $L_{\mathrm{L}}^{\mathrm{mappo}}\left( \vartheta \right)$ via mini-batch SGA.

\item We then fix the low-level policy $\pi_{\mathrm{L}}(p_{m,t}|S_{m,t},\omega_{t},m;\vartheta)$ and update the Lagrangian multiplier $\lambda_{\mathrm{L}}$ according to
\begin{align}
\label{update-lambdaL}
    \lambda _{\mathrm{L}}\gets \mathrm{ReLU}\left( \lambda _{\mathrm{L}}-\alpha _{\lambda}\cdot \nabla _{\lambda _{\mathrm{L}}}\left( \mathcal{L} _{\mathrm{L}}\left( \pi _{\mathrm{L}},\lambda _{\mathrm{L}} \right) \right) \right) \\ \IEEEnonumber
    =\mathrm{ReLU}\left( \lambda _{\mathrm{L}}+\alpha _{\lambda}\cdot \left( J_{\mathrm{L}}^{\mathrm{safe}}\left( \pi _{\mathrm{L}} \right) -d \right) \right).  
\end{align}

\item Finally, we update the two critic networks by optimizing the following objective functions, i.e.,
\begin{equation}
L^{\mathrm{critic}}_{\mathrm{R}}\left( \phi _{\mathrm{R}} \right) =\frac{1}{ET}\sum_{e=1}^E\sum_{t=0}^{T-1}  \left( V_{\mathrm{R}}\left( S_t,\omega_t;{\phi _{\mathrm{R}}} \right) -{R}_{t}^{\mathrm{}} \right) ^2 , 
\label{loss-critic-R}
\end{equation}
\begin{equation}
L^{\mathrm{critic}}_{\mathrm{C}}\left( \phi _{\mathrm{C}} \right) =\frac{1}{ET}\sum_{e=1}^E\sum_{t=0}^{T-1} \left( V_{\mathrm{C}}\left( S_t,\omega_t ;{\phi _{\mathrm{C}}}\right) -{C}_{t}^{\mathrm{}} \right) ^2 ,
\label{loss-critic-C}
\end{equation}
\noindent where the returns ${R}_{t}^{\mathrm{}}$ and ${C}_{t}^{\mathrm{}}$ are given by \eqref{return-opr} and \eqref{return-safe}, respectively. The parameters $\phi_{\mathrm{R}}$ and $\phi_{\mathrm{C}}$ are updated to minimize $L^{\mathrm{critic}}_{\mathrm{R}}\left( \phi _{\mathrm{R}} \right)$ and $L^{\mathrm{critic}}_{\mathrm{C}}\left( \phi _{\mathrm{C}} \right)$ via mini-batch stochastic gradient
descent (SGD), respectively.

\end{itemize}

The pseudocode of DAC-MAPDO is detailed in Algorithm \ref{alg-DAC-MAPPO}. For further technical details, the optimization procedures for PPO and MAPPO are referenced in \cite{chen2018adaptive} and \cite{yu2022surprising}, respectively.

{\color{black}
\subsection{Duality Gap Analysis}
Since the DAC-MAPPO-Lagrangian algorithm is based on primal dual method and the optimization objective $\mathbf{P}$ in \eqref{cobjective} is not convex, we establish the convergence to the optimal solution by theoretically proving that the duality gap is approximately zero.  \par

As the primal dual method is employed in both the High-Level and Low-Level CMDPs, we confine our analysis to the duality gap of the High-Level CMDP, with analogous results holding for the Low-Level CMDP.\par

Let $\mathcal{P}(\mathcal{S}_{\rm H})$ denote the space of probability measures on $(\varOmega_t,\hat{\mathcal{B}}(\varOmega_t))$, indexed by elements of $\mathcal{S}_{\rm H}$, where $\hat{\mathcal{B}}(\varOmega_t)$ denotes the Borel $\sigma$-algebra of $\varOmega_t$. The agent chooses options sequentially based on the high-level policy $\pi_{\rm H}\in\mathcal{P}(\mathcal{S}_{\rm H})$. As a widely used class of parameterizations, DNNs are able to model any function in $\mathcal{P}(\mathcal{S}_{\rm H})$ to within a stated accuracy. We formalize this concept in the following definition.

\begin{mydef}
A parametrization $\pi_{\rm H}(\cdot;\theta,\varphi)$ is an $\epsilon$-universal parameterization of functions in $\mathcal{P}(\mathcal{S}_{\rm H})$ if, for some $\epsilon> 0$, there exists for any $\pi_{\rm H}\in\mathcal{P}(\mathcal{S}_{\rm H})$ parameters ${\theta,\varphi}\in\mathbb{R}^p$ such that
\begin{align}
\label{episilon}
    \underset{s\in \mathcal{S}_{\rm H}}{\max}\sum_{\omega\in \varOmega_t}^{}{\left| \pi_{\rm H} \left( \omega|s \right) -\pi _{\rm H}\left( \omega|s;\theta,\varphi \right) \right|\leqslant \epsilon}.
\end{align}
\end{mydef}

\noindent Before presenting Theorem 1, we first define the \emph{perturbation function} associated with problem \eqref{cobjectivehigh}. For any $\xi\in \mathbb{R}^{5m+3}$ (where $5m+3$ is the dimension of the state space), the perturbation function associated with \eqref{cobjectivehigh} is defined as
\begin{equation}
    \begin{aligned}
\label{xi1}
    &\mathbf{P}_{\rm H}(\xi)=\underset{\pi_{\rm H}\in \mathcal{P}(\mathcal{S}_{\rm H})}{\max}\,J_{\rm H}^{\rm opr}(\pi_{\rm H}),\\
    \mathrm{s.t.} \quad &  J_{\rm H}^{\mathrm{safe}}(\pi_{\rm H})\leq d+\xi. 
\end{aligned}
\end{equation}

Notice that $\mathbf{P}_{\rm H}(0)=\mathbf{P}_{\rm H}^{*}$, which corresponds to the optimal value of problem \eqref{cobjectivehigh}.

\begin{theorem2}
\label{dualgap}
Suppose that Slater’s condition holds for \eqref{cobjectivehigh}. For the optimal value $\mathbf{P}_{\rm H}^*$ of the original problem in \eqref{cobjectivehigh} and the optimal value $\mathbf{D}_{\rm H,\theta,\varphi}^{*}$ of the parametrized Lagrange dual problem in \eqref{minmax_high}, it follows that
\begin{align}
\label{parametergap}
    \mathbf{P}_{\rm H}^*\geqslant \mathbf{D}_{\rm H,{\theta,\varphi}}^{*}\geqslant \mathbf{P}_{\rm H}^*-\left( B_{r}+\left\| \lambda _{\epsilon}^{*} \right\| _1B_c \right) \frac{\epsilon}{1-\gamma},
\end{align} 
\noindent where $B_{r}>0$ and $B_c>0$ are the upper bounds for $r_{\rm H}^{\rm opr}(s,\omega)$ and $-c_{\rm H}^{\rm safe}(s)$, respectively. 
$\gamma$ is a discount factor, and $\epsilon$ is given in \eqref{episilon}. Let $\lambda _{\epsilon}^{*} $ be the solution to the dual problem associated with \eqref{xi1} for perturbation $\xi=B_c\epsilon/(1-\gamma)$. 
\end{theorem2}

The proof of Theorem \ref{dualgap} is in the Appendix B. Theorem \ref{dualgap} implies that by solving the dual problem $\mathbf{D}_{\rm H,{\theta},\varphi}^{*}$ for the high-level CMDP, the resulting sub-optimality is of order $\epsilon$, which corresponds to the error in representing the policies. By improving the representation capability of the DNNs, for example by increasing its size, this error can be made arbitrarily small. When the representation error $\epsilon$ is zero, the duality gap also vanishes, meaning that solving the original constrained problem $\mathbf{P}_{\rm H}^*$ and the dual problem $\mathbf{D}_{\rm H,{\theta},\varphi}^{*}$ are equivalent. Finally, assuming that Slater’s condition holds for \eqref{cobjectivehigh} implies the existence of at least one policy that satisfies the constraint in \eqref{cobjectivehigh} with strict inequality. This assumption is typically mild and easily met in practice. 

}

\section{Numerical Analysis}
In this section, we conduct experiments to evaluate the effectiveness of the proposed algorithm based on real-world data. All the experiments are performed on a Linux server, using Python 3.8 with Pytorch to implement the DRL-based approaches. 

\subsection{Experimental Setup}
\subsubsection{Experimental data}
\begin{figure}[t]
\centering
\includegraphics[width=0.95\linewidth]{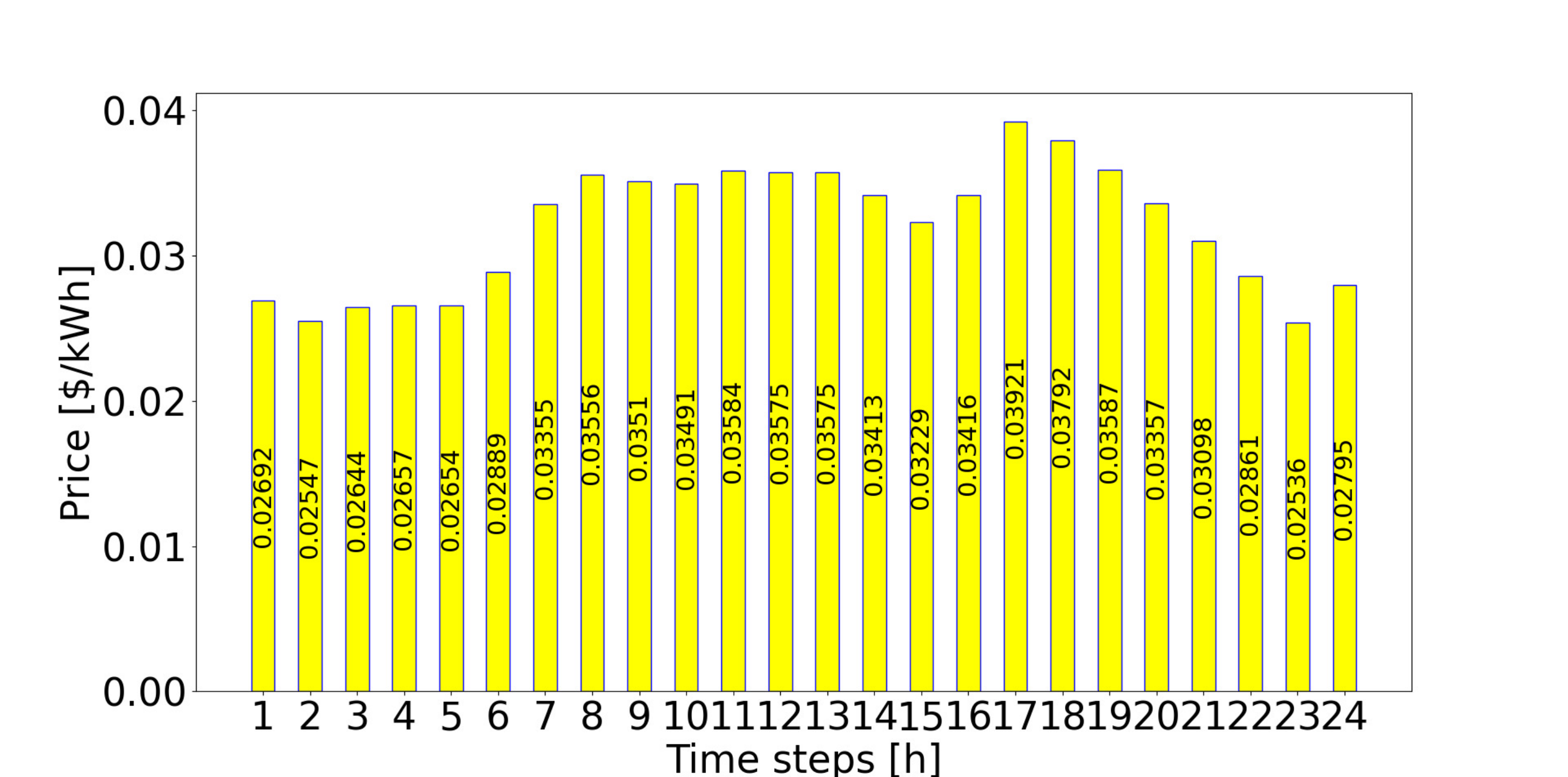}
\caption{The variation of the electricity prices of a typical day in the experimental data. }
\label{dataset} 
\end{figure}

We utilize real-world datasets on grid electricity prices and PV output power to train and evaluate the proposed DRL-based algorithm. The dataset of time-varying electricity prices is obtained from the Midcontinent Independent System Operator (MISO) \cite{MISO2023}. Fig. \ref{dataset} illustrates the hourly electricity price profile on a typical day. It can be observed that electricity prices exhibit significant temporal fluctuations, with the peak price reaching $\$0.03921$/kWh during the evening demand surge between 17:00 and 18:00.

\begin{figure}[t]
\centering
\includegraphics[width=0.95\linewidth]{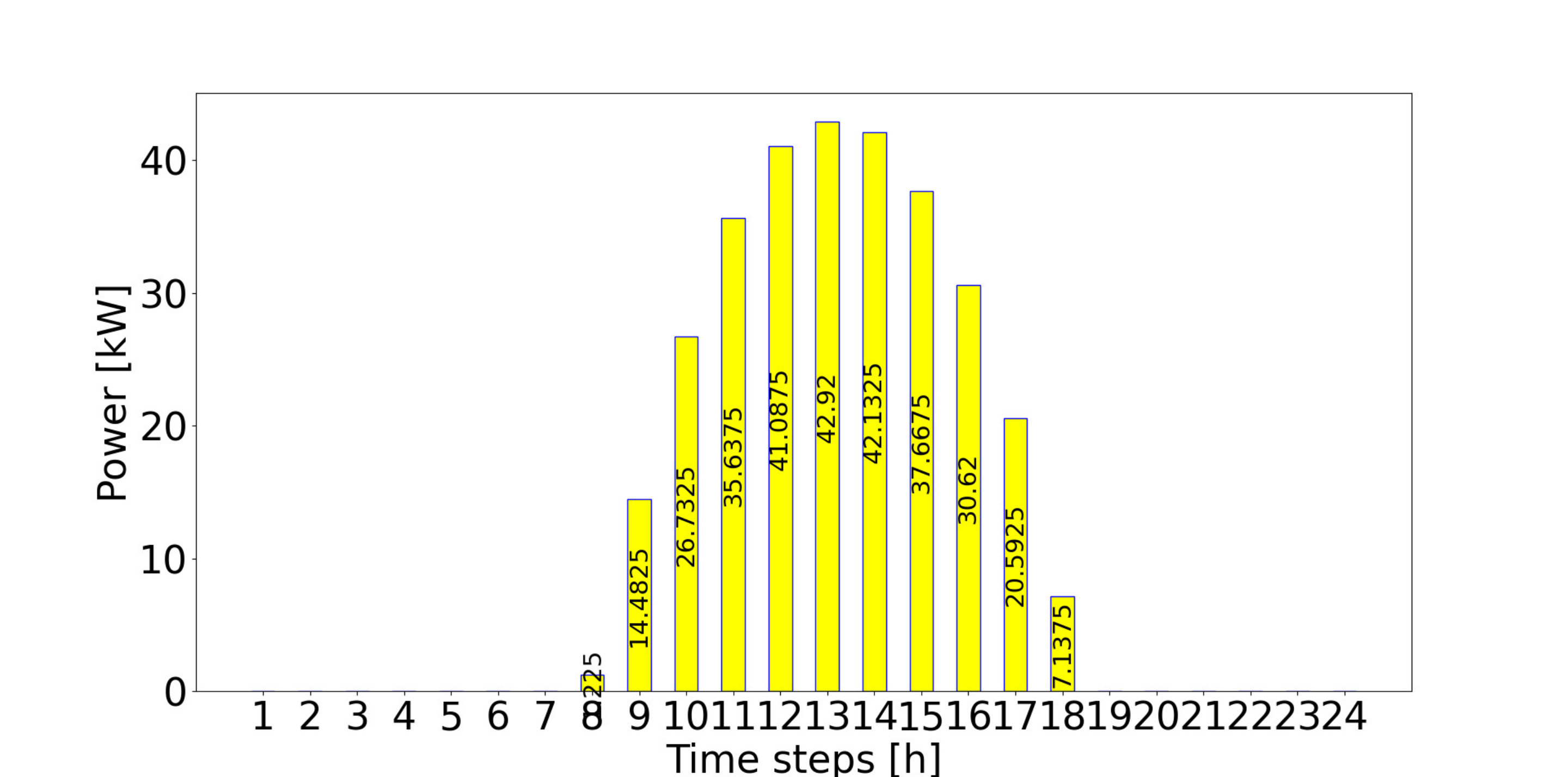}
\caption{The variation of the PV power of a typical day in the experimental data. }
\label{datasetPV} 
\end{figure}

The PV power dataset is collected from an isolated microgrid \cite{9067138}. Fig. \ref{datasetPV} shows the hourly PV power in a typical day, which peaks at $42.92$ kW at 13:00. Notably, PV generation remains nearly zero during the early morning and evening periods (i.e., before 8:00 and after 18:00), which coincides with the hours of high electricity prices. This temporal mismatch between high electricity prices and low PV power enables the DRL agent to learn intelligent charging and discharging policies. By leveraging PV generation when available and avoiding grid power during peak price periods, the agent can effectively minimize costs while satisfying operational constraints.\par

\begin{figure}[t]
\centering
\includegraphics[width=0.8\linewidth]{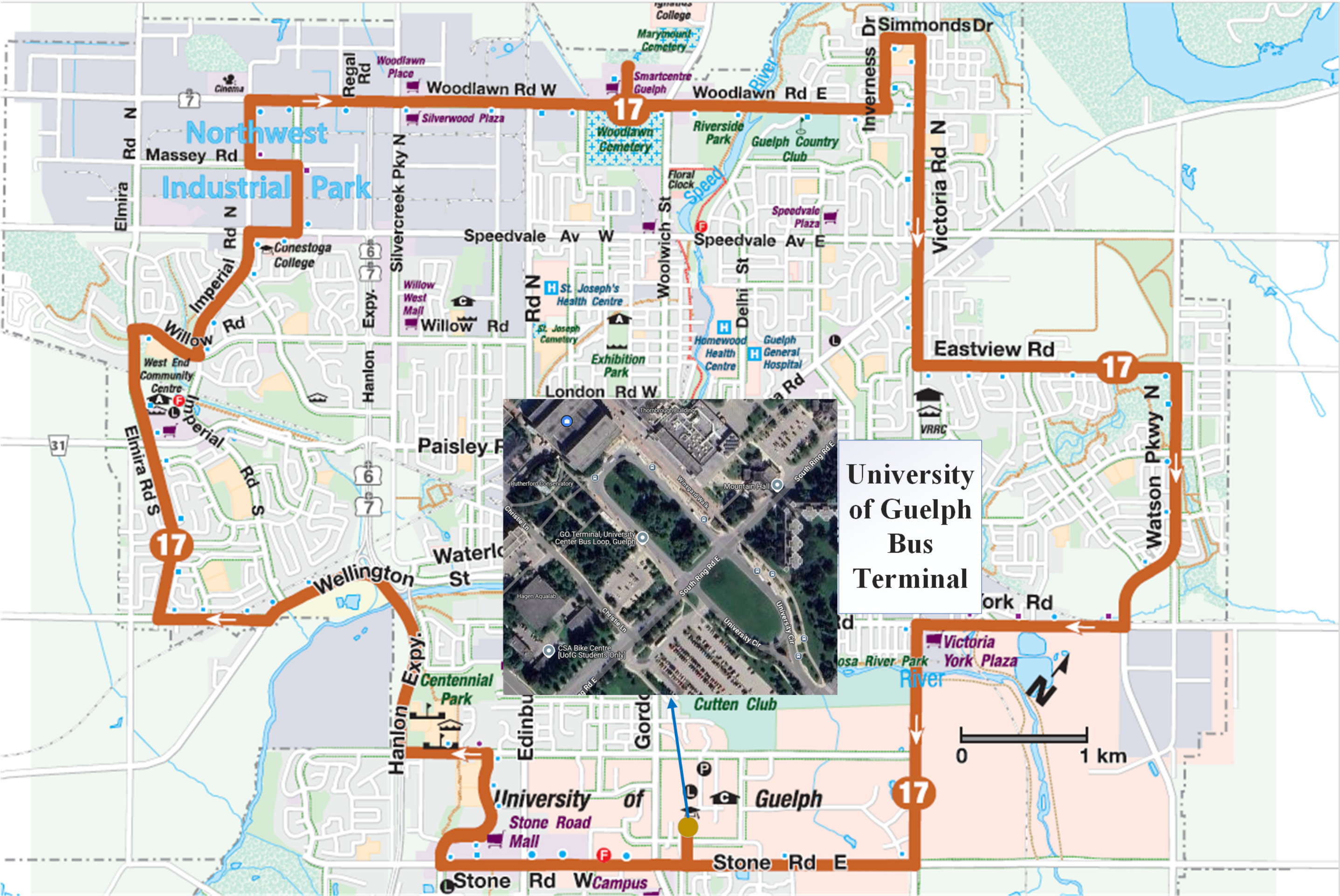}
\caption{An illustration of the bus network and terminal station used in this study. }
\label{terminal-map} 
\end{figure}

We consider that the sustainable terminal station is situated on the University of Guelph campus. The bus schedule used in our experiments follows a predefined timetable provided by Guelph Transit \cite{busschedule}. Fig. \ref{terminal-map} illustrates the bus network and the sustainable terminal station used in this study.

\subsubsection{Simulated Environment}
  

{\color{black}To evaluate the performance and scalability of the proposed framework, two experimental scenarios are considered:
\begin{itemize}
\item \textbf{Scenario 1}: A smaller fleet with $M=6$ EBs and $N=3$ chargers, where four PV panels are installed at the terminal station.
\item \textbf{Scenario 2}: A larger fleet with $M=20$ EBs and $N=10$ chargers, with PV generation capacity scaled proportionally to thirteen panels at the terminal station.
\end{itemize}}
  
All EBs operate on $L=2$ bus routes, namely Route 17 and Route 18, which follow the same path in opposite directions, forming a pair of loop lines as illustrated in Fig.~\ref{terminal-map}. \par

 
  The operating duration for each EB is modeled as a stochastic variable following two normal distributions: $\mathcal{N}(50,8)$ minutes during peak traffic hours (7:00–9:00 AM and 5:00–7:00 PM), and $\mathcal{N}(40,8)$ minutes during non-peak traffic hours. The time step duration is set to $\Delta t=10 \ \mathrm{min}$. \par 
  
  The environment parameters, summarized in Table~\ref{parameters}, are identical for both scenarios. {\color{black}The minimum SoC level $\eta_{\mathrm{min}}=0.2$ follows existing works \cite{liu2024electric,zaneti2022sustainable}, where the 20\% reserve serves as a safety buffer to prevent full depletion, thereby ensuring reliable operation even under increased seasonal loads, such as heating, ventilation, and air conditioning (HVAC) usage during peak summer or winter conditions.}\par 

  {\color{black}The experiments were conducted on a workstation with $32$ GB RAM, an $8$-core CPU, and an NVIDIA A2000 GPU. The training time was approximately $30$ hours for the $6$-EB, $3$-charger scenario and $46$ hours for the $20$-EB, $10$-charger scenario. Once trained, the policies are lightweight: during execution, each decision step requires less than one second for inference, with minimal memory usage. Therefore, the proposed framework is suitable for deployment on edge hardware at bus depots to support real-time scheduling.}
  
\subsubsection{Baseline algorithms}
We compare the performance of the proposed DAC-MAPPO-Lagrangian algorithm with the following three baseline methods.
\begin{itemize}
    \item {\color{black}\emph{MILP under Deterministic setting (MILP-D): A benchmark optimization model formulated as an MILP problem, where all parameters are assumed to be perfectly known in advance. The optimization objectives and constraints are the same as those specified in Section II.D. In this deterministic scenario, the electricity prices, PV generation power, and travel times of all EBs are fully known, allowing the MILP to be solved using a commercial optimization solver. This solution serves as the oracle and provides a benchmark for the theoretical upper bound of system performance, as it operates with perfect information.}}
    \item \emph{MILP under Stochastic setting (MILP-S)}: A practical extension of MILP-D that accounts for uncertainty in system parameters. To enable the MILP formulation under the stochastic setting of our system model—where electricity prices, PV generation power, and travel times are treated as uncertain variables—travel times are estimated using K-means clustering on historical data, based on features such as EB ID, trip ID, and departure time, following a similar approach in \cite{he2023battery}. For electricity prices, the day is divided into six intervals: (0:00–6:00), (6:00–9:00), (9:00–14:00), (14:00–17:00), (17:00–21:00), and (21:00–24:00), with fixed values assigned to each interval \cite{liu2024electric}. These values are estimated from the average historical data for the same periods in the week preceding the test day.


\item \emph{DAC-MAPPO \cite{10930901}}: A DRL-based algorithm similar to the proposed approach, but designated to solve the original MDP formulation presented in Section~III.A, rather than the CMDP. The battery depletion penalty is incorporated into the reward function as a negative term weighted by a fixed coefficient $\lambda^\mathrm{safe}$, whose value is determined via grid search and listed in Table \ref{parameters}. As a model-free DRL algorithm, this method can directly handle uncertainties without requiring forecasts of electricity prices or PV generation. Therefore, it uses the same real-time datasets as the proposed algorithm.
 
\end{itemize}

 The hyperparameters of the DRL algorithms are listed in Table \ref{DRLparameters}, {\color{black}selected based on prior literature and further tuned through grid search within candidate ranges to ensure stable training and robust performance.} Since the high-level actor network is centralized, we adopt a larger network size to enable learning of a more complex policy over options. The tolerance value for safety constraint violation, $d$, is set to $0.025$, {\color{black}chosen through grid search to balance operational cost and violation rate, ensuring both safe and cost-efficient scheduling performance.}\par
 

\begin{table}[t]
\centering
\caption{Parameter configuration in the system model.}
\begin{tabular}[b]{p{1.5cm}<{\raggedright}p{0.9cm}<{\raggedright}p{5.0cm}<{\raggedright}}
\hline
\textbf{Notations}&\textbf{Values}&\textbf{Description}\\
\hline
\specialrule{0em}{1pt}{1pt}
$h$ & 4 &The length of the time window to look into the historical PV power and prices \\
\specialrule{0em}{1pt}{1pt}
$p^{\rm ch}_{\max}$/ $p^{\rm dis}_{\max}$ & 120kW/ 120kW &The maximum absolute value of charging/discharging power of EBs\\
\specialrule{0em}{1pt}{1pt}
$\eta_{\mathrm{min}}$ / $\eta_{\mathrm{max}}$    & \textcolor{black}{0.2} / 1 &The minimum/maximum SoC level\\
\specialrule{0em}{1pt}{1pt}
$E_{\mathrm{bat}}$ & 240kWh &The battery capacity\\
\specialrule{0em}{1pt}{1pt}
$\lambda^\mathrm{safe}$    & 2.5 &The weight of battery depletion penalty in MILP-S and DAC-MAPPO\\
\specialrule{0em}{1pt}{1pt}
$\zeta^\mathrm{b}$    & 0.1 &The weight of the battery degradation cost\\
\specialrule{0em}{1pt}{1pt}
$\zeta^\mathrm{s}$    & 0.1 &The constant value used to quantify the switching cost\\
\specialrule{0em}{1pt}{1pt}

\specialrule{0.05em}{2pt}{0pt}
\end{tabular}
\label{parameters}
\end{table}

\begin{table}[t]
\centering
\caption{Hyperparameters of DRL-based algorithms.}
\begin{tabular}[b]{p{1.9cm}<{\centering}p{1.4cm}<{\centering}p{2.1cm}<{\centering}p{1.2cm}<{\centering}}
\hline
\textbf{Networks}&\textbf{High-level}&\textbf{Low-level}& \textbf{Learning Rate}   \\
\specialrule{0.05em}{0pt}{0pt}

\specialrule{0em}{1pt}{1pt}
Actor   &128,128 &64,64  &3e-4   \\
\specialrule{0em}{1pt}{1pt}
Termination condition    &64,64&\textbackslash{} & 3e-4   \\		
\specialrule{0em}{1pt}{1pt}
\hline
\specialrule{0em}{1pt}{1pt}
Reward critic    &\textbackslash{}&128,128 & 1e-3   \\		
\specialrule{0em}{1pt}{1pt}
Cost critic    &\textbackslash{}&128,128 & 1e-3   \\		
\specialrule{0em}{1pt}{1pt}


\specialrule{0em}{1pt}{1pt}
\hline
\specialrule{0em}{1pt}{1pt}
\specialrule{0.05em}{3pt}{3pt}
\specialrule{0em}{1pt}{1pt}
\textbf{Parameters}&\textbf{Value}&\textbf{Parameters}& \textbf{Value} \\
\specialrule{0em}{1pt}{1pt}
\hline
\specialrule{0em}{1pt}{1pt}
\multicolumn{1}{l}{GAE coefficient}&0.95&Clip $ \varepsilon$&0.2 \\
\specialrule{0em}{1pt}{1pt}
Iterations&20,000&Episodes $E$  & 10 \\
\specialrule{0em}{1pt}{1pt}
 Batch size & 128 &\textcolor{black}{Learning rate $\alpha_{\lambda}$} &\textcolor{black}{0.01}\\
 \specialrule{0em}{1pt}{1pt}
 Tolerance $d$& 0.025 & &\\
 \specialrule{0em}{1pt}{1pt}
\hline
\end{tabular}
\label{DRLparameters}
\end{table}

\subsubsection{Performance Metrics}
To evaluate the performance of the algorithms, inspired by \cite{8910361,feng2024safe}, we focus on two primary metrics:
\begin{itemize}
    \item \emph{Average operational return}, defined in \eqref{opr-return}, which is the average cumulative operational reward across all test episodes. The operational reward - equivalently, the negative operational cost $c_{t}^{\mathrm{opr}}$ (see \eqref{opr-reward}) - comprises the charging cost $c_{t}^{\mathrm{ch}}$, the battery cost $c_{m,t}^{\mathrm{bat}}$, and the switching cost $c_{m,t}^{\mathrm{sw}}$ as detailed in \eqref{oprcost}.
    \item \emph{Safety violation rate}, which is defined as the percentage of considered episodes in which at least one safety violation occurs, i.e., where the safety cost $c_{t}^{\mathrm{safe}}$ in \eqref{Costend} is non-zero at any time step.
\end{itemize}

\subsection{Experimental Results}

\subsubsection{Performance for the test set}
We use one year of 2022 data for training and conduct three independent runs. For the evaluation, the 2023 data are divided into three four-month periods, i.e., January–April, May–August, and September–December, with each run evaluated on one of the periods. Given the strong seasonal variability in both electricity prices and PV generation, this evaluation method allows us to assess the performance of the learned policy across different seasons. In each run, we evaluate the trained policy over $500$ complete test episodes and compute the average operational return and safety violation rate over these episodes. \par


Table~\ref{result} summarizes the average operational return and safety violation rate for each individual run, along with the overall mean and standard deviation across the three runs {\color{black}for both Scenario 1 and Scenario 2.} \par 

\begin{table*}[ht]
\centering
\caption{The individual and mean performances, as well as the standard deviation of all the algorithms across three independent runs \textcolor{black}{under two experimental scenarios}.}
\label{result}
{\color{black}
\begin{tabular}{|c|c|l|ccccc|}
\hline
\multirow{2}{*}{\textbf{Scenarios}} & \multirow{2}{*}{\textbf{\begin{tabular}[c]{@{}c@{}}Performance \\ Metrics\end{tabular}}}                                                               & \multirow{2}{*}{\textbf{Algorithms}} & \multicolumn{5}{c|}{\textbf{Results}}                                                            \\ \cline{4-8} 
                                   &     &                                      & \multicolumn{1}{c|}{\textbf{Run 1}} & \multicolumn{1}{c|}{\textbf{Run 2}} & \multicolumn{1}{c|}{\textbf{Run 3}} & \multicolumn{1}{c|}{\textbf{Mean}} & \textbf{\begin{tabular}[c]{@{}c@{}}Standard \\ Deviation\end{tabular}} \\ \hline
\multirow{8}{*}{\textbf{1}}        & \multirow{4}{*}{\textbf{\begin{tabular}[c]{@{}c@{}}Average \\ Operational \\ Return\end{tabular}}} & MILP-D                              & \multicolumn{1}{c|}{-12.54}         & \multicolumn{1}{c|}{-12.59}         & \multicolumn{1}{c|}{-12.51}         & \multicolumn{1}{c|}{-12.55}           & 0.040                                                                 \\ \cline{3-8} 
 &                      & MILP-S                            & \multicolumn{1}{c|}{-13.69}         & \multicolumn{1}{c|}{-14.30}         & \multicolumn{1}{c|}{-13.76}         & \multicolumn{1}{c|}{-13.92}           & 0.334                                                                   \\ \cline{3-8} 
                                   &                      & DAC-MAPPO                            & \multicolumn{1}{c|}{-12.85}         & \multicolumn{1}{c|}{-12.93}         & \multicolumn{1}{c|}{-12.77}         & \multicolumn{1}{c|}{-12.85}           & 0.080                                                                   \\ \cline{3-8} 
                                   &                                                                                                     & DAC-MAPPO-Lagrangian                 & \multicolumn{1}{c|}{-12.61}         & \multicolumn{1}{c|}{-12.69}         & \multicolumn{1}{c|}{-12.58}         & \multicolumn{1}{c|}{-12.63}           & 0.057                                                                   \\ \cline{2-8} 
                                   & \multirow{4}{*}{\textbf{\begin{tabular}[c]{@{}c@{}}Safety \\ Violation \\ Rate\end{tabular}}}       & MILP-D                               & \multicolumn{1}{c|}{0.0\%}           & \multicolumn{1}{c|}{0.0\%}           & \multicolumn{1}{c|}{0.0\%}           & \multicolumn{1}{c|}{0.0\%}           & 0.000                                                                 \\ \cline{3-8}
                                   &   & MILP-S                            & \multicolumn{1}{c|}{15.8\%}           & \multicolumn{1}{c|}{18.6\%}           & \multicolumn{1}{c|}{16.0\%}           & \multicolumn{1}{c|}{16.8\%}           & 0.016             \\ \cline{3-8}
                                   &   & DAC-MAPPO                            & \multicolumn{1}{c|}{8.6\%}           & \multicolumn{1}{c|}{12.6\%}           & \multicolumn{1}{c|}{5.4\%}            & \multicolumn{1}{c|}{8.9\%}           & 0.036             \\ \cline{3-8} 
                                   &                                                                                                     & DAC-MAPPO-Lagrangian                 & \multicolumn{1}{c|}{0.0\%}            & \multicolumn{1}{c|}{2.4\%}            & \multicolumn{1}{c|}{0.0\%}            & \multicolumn{1}{c|}{0.8\%}            & 0.014                                                                  \\ \hline
\multirow{8}{*}{\textbf{2}}        & \multirow{4}{*}{\textbf{\begin{tabular}[c]{@{}c@{}}Average \\ Operational \\ Return\end{tabular}}} & MILP-D                               & \multicolumn{1}{c|}{-45.69}         & \multicolumn{1}{c|}{-45.98}         & \multicolumn{1}{c|}{-45.62}         & \multicolumn{1}{c|}{-45.76}           & 0.191                                                               \\ \cline{3-8} 
 &   & MILP-S                           & \multicolumn{1}{c|}{-50.85}         & \multicolumn{1}{c|}{-52.85}         & \multicolumn{1}{c|}{-49.70}         & \multicolumn{1}{c|}{-51.13}           & 1.594     \\ \cline{3-8}
                                   &   & DAC-MAPPO                            & \multicolumn{1}{c|}{-48.76}         & \multicolumn{1}{c|}{-52.83}         & \multicolumn{1}{c|}{-47.37}         & \multicolumn{1}{c|}{-49.65}           & 2.838                                                                   \\ \cline{3-8} 
                                   &                                                                                                     & DAC-MAPPO-Lagrangian                 & \multicolumn{1}{c|}{-45.81}         & \multicolumn{1}{c|}{-46.74}         & \multicolumn{1}{c|}{-45.79}         & \multicolumn{1}{c|}{-46.11}           & 0.543                                                                   \\ \cline{2-8} 
                                   & \multirow{4}{*}{\textbf{\begin{tabular}[c]{@{}c@{}}Safety \\ Violation \\ Rate\end{tabular}}}       & MILP-D                               & \multicolumn{1}{c|}{0.0\%}           & \multicolumn{1}{c|}{0.0\%}           & \multicolumn{1}{c|}{0.0\%}           & \multicolumn{1}{c|}{0.0\%}           & 0.000                                                                  \\ \cline{3-8} 
                                    &    & MILP-S                 & \multicolumn{1}{c|}{16.6\%}           & \multicolumn{1}{c|}{20.8\%}           & \multicolumn{1}{c|}{14.8\%}           & \multicolumn{1}{c|}{17.4\%}           & 0.031                                                                  \\ \cline{3-8}
                                   &                                                                                                     & DAC-MAPPO                            & \multicolumn{1}{c|}{12.8\%}           & \multicolumn{1}{c|}{20.4\%}           & \multicolumn{1}{c|}{5.8\%}            & \multicolumn{1}{c|}{13.0\%}           & 0.073                                                                  \\ \cline{3-8} 
                                   &    & DAC-MAPPO-Lagrangian                 & \multicolumn{1}{c|}{0.2\%}            & \multicolumn{1}{c|}{3.2\%}            & \multicolumn{1}{c|}{0.0\%}            & \multicolumn{1}{c|}{1.1\%}            & 0.018                                                                  \\ \hline
\end{tabular}
}
\end{table*}

The performance rankings of all algorithms remain consistent across both scenarios. {\color{black}MILP-D, representing the theoretical optimal solution, achieved the best average performance. The proposed DAC-MAPPO-Lagrangian algorithm closely followed. In Scenario 1, DAC-MAPPO-Lagrangian achieved the average operational return of $-12.63$, only $0.64$\% lower than MILP-D, with the standard deviation of $0.057$. In Scenario 2, it again outperformed DAC-MAPPO and MILP-S, achieving $-46.11$, only $0.96$\% lower than MILP-D, with a standard deviation of $0.018$. DAC-MAPPO ranked third in both scenarios ($-12.85$ and $-49.65$), showing higher standard deviations ($0.080$ and $0.073$), while MILP-S consistently yielded the lowest returns, lagging behind DAC-MAPPO-Lagrangian by $10.21\%$ and $10.89\%$ in Scenarios 1 and 2, respectively. In addition, the improvement of the Lagrangian variant over the original DAC-MAPPO grows from $1.74\%$ to $7.68\%$ as the fleet size increases, indicating enhanced scalability and stability.}\par

Regarding safety performance, {\color{black}as the theoretical optimal solution, MILP-D achieved zero violation rates for all runs in both scenarios.} DAC-MAPPO-Lagrangian achieved low average violation rates{\color{black}—$0.8\%$ and $1.1\%$ in Scenarios 1 and 2—while DAC-MAPPO recorded $8.9\%$ and $13.0\%$, and MILP-S performed worst with $16.8\%$ and $17.4\%$.} Notably, DAC-MAPPO-Lagrangian achieved zero violations in two of the three runs in Scenario 1, confirming strong constraint satisfaction. {\color{black}Overall, across both fleet scales, DAC-MAPPO-Lagrangian consistently achieved the best operational return and the lowest violation rate among all algorithms (except MILP-D), with superior scalability as system size increases.}


The fact that both DAC-MAPPO and DAC-MAPPO-Lagrangian significantly outperform MILP-S highlights the strength of DRL-based approaches in handling uncertainty. The primary limitation of MILP-S lies in its lack of adaptability, as it relies on forecasted values of travel time, PV output, and electricity prices. When actual conditions deviate from these forecasts, the precomputed decisions may become suboptimal. In contrast, as model-free DRL methods, DAC-MAPPO and DAC-MAPPO-Lagrangian continuously adjust their decisions based on real-time state observations, making them inherently more suitable for dynamic and uncertain environments. {\color{black}This performance gap becomes even more pronounced in Scenario 2, where the enlarged state and action spaces amplify uncertainty.} 

{\color{black}To further analyze the effect of forecast errors, we compared MILP-S with MILP-D in Run 1, Scenario 1. MILP-S suffers approximately $9.17$\% degradation in average operational return relative to MILP-D due to forecast inaccuracies—primarily an average $13.9$\% PV generation underestimation and $5.1$\% electricity price underestimation. In contrast, DAC-MAPPO-Lagrangian, which makes decisions based on real-time state observations, achieves average operational returns close to the deterministic theoretical optimal solution while maintaining strong safety compliance, demonstrating its superior adaptability and robustness under uncertainty.}\par

Next, we observe that although both DAC-MAPPO and DAC-MAPPO-Lagrangian achieve comparable and favorable average operational returns, the DAC-MAPPO algorithm exhibits instability in terms of safety performance. Taking Scenario 1 as an example, while DAC-MAPPO achieves a low violation rate of only $5.4$\% in Run 3, the rate spikes to $12.6$\% in Run 2. This inconsistency arises from the multi-objective nature of the EBCSP, which requires balancing the minimization of operational costs with the enforcement of safety. In DAC-MAPPO, the reward function comprises two components: one for cost efficiency and the other for safety, and the results are highly sensitive to the manually chosen weight $\lambda^{\rm safe}$. Meanwhile, the DAC-MAPPO-Lagrangian algorithm explicitly models safety as a constraint and addresses the resulting CMDP using primal–dual optimization. It adaptively adjusts the weights of the safety cost through gradient ascent on the Lagrange multipliers $\lambda_{\rm H}$ and $\lambda_{\rm L}$. {\color{black}This adaptive mechanism proves even more advantageous in the large-scale Scenario 2, where the expanded agent interactions make manual tuning increasingly difficult. The Lagrangian formulation enables stable convergence and consistent safety compliance across both scenarios, highlighting its scalability and robustness.

}\par

\subsubsection{Convergence Properties}

\begin{figure*}[t]
\centering
\subfigure[S1: Average operational return]{
\label{training-returnS1} 
\includegraphics[width=0.22\textwidth]{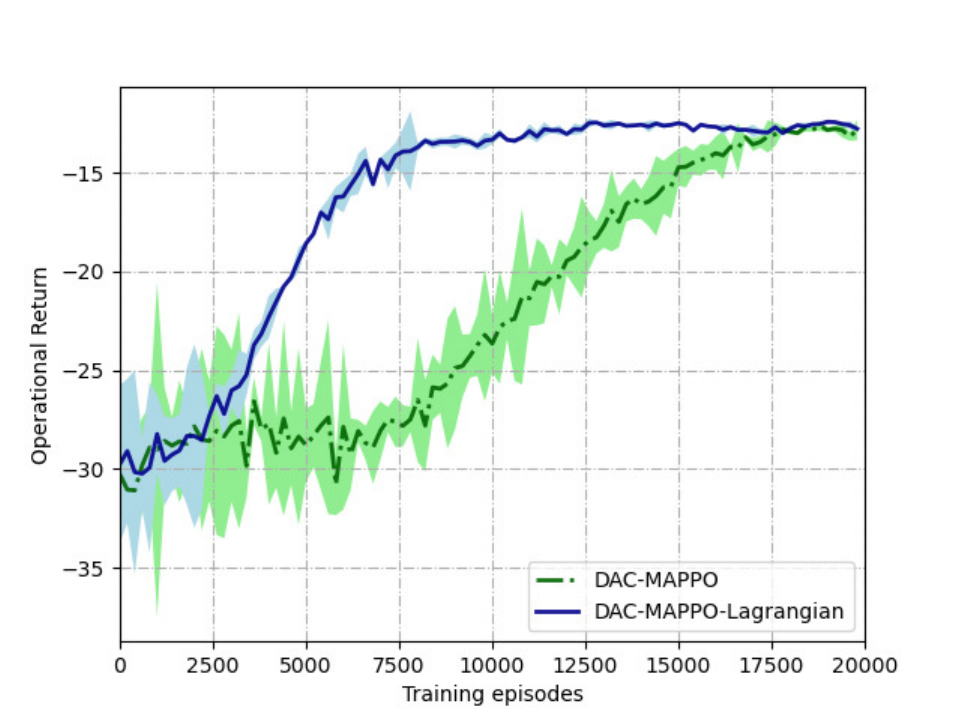}}
\subfigure[S1: Safety violation rate]{
\label{training-rateS1} 
\includegraphics[width=0.22\textwidth]{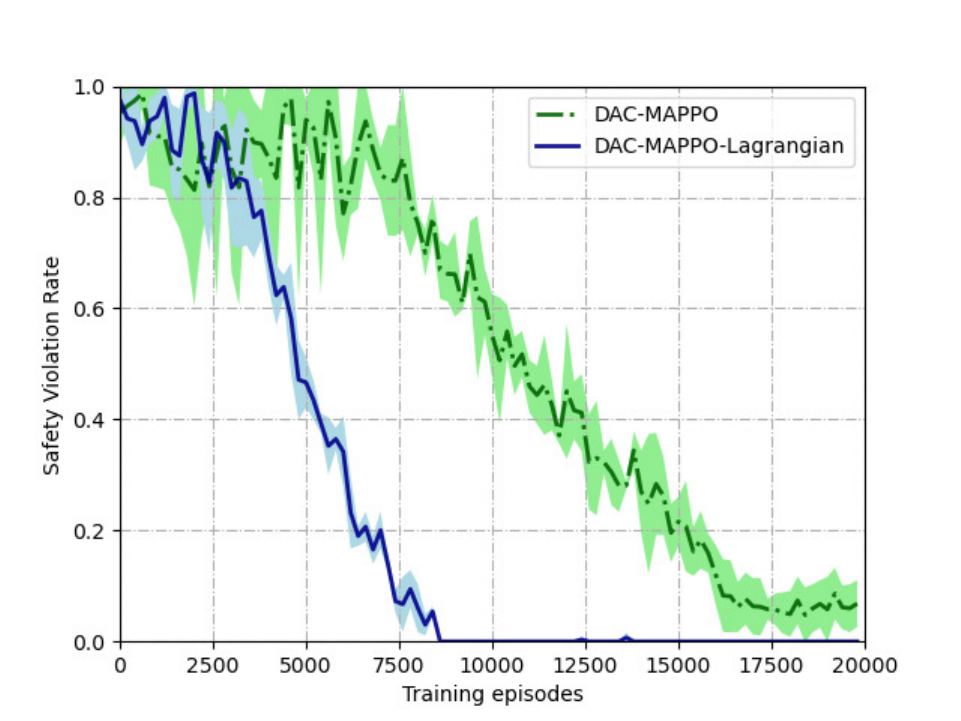}}
\subfigure[S2: Average operational return]{
\label{training-returnS2} 
\includegraphics[width=0.22\textwidth]{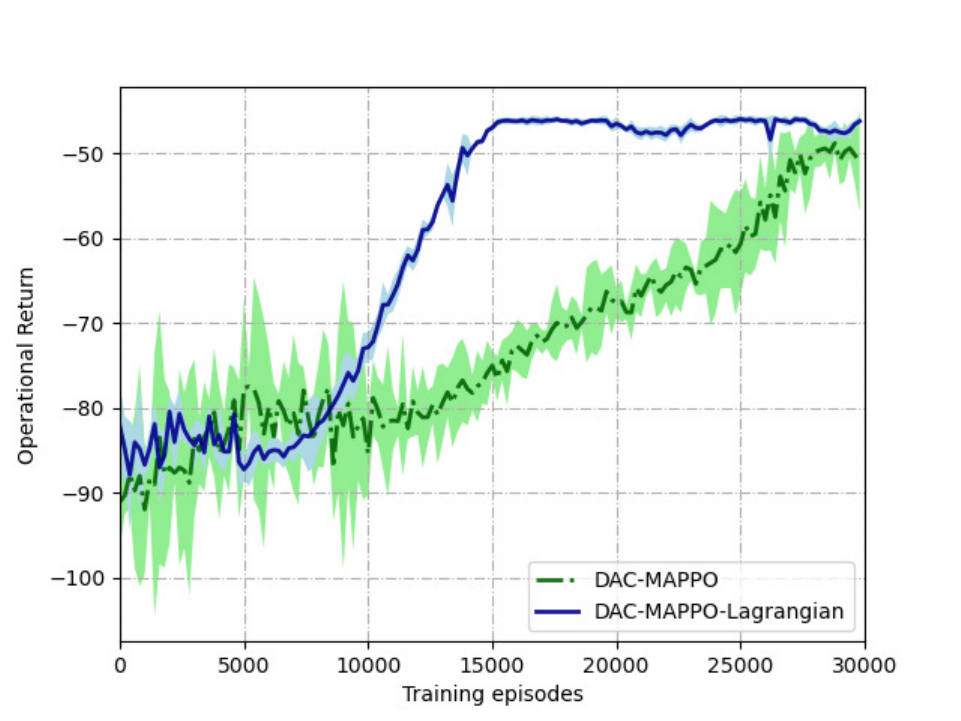}}
\subfigure[S2: Safety violation rate]{
\label{training-rateS2} 
\includegraphics[width=0.22\textwidth]{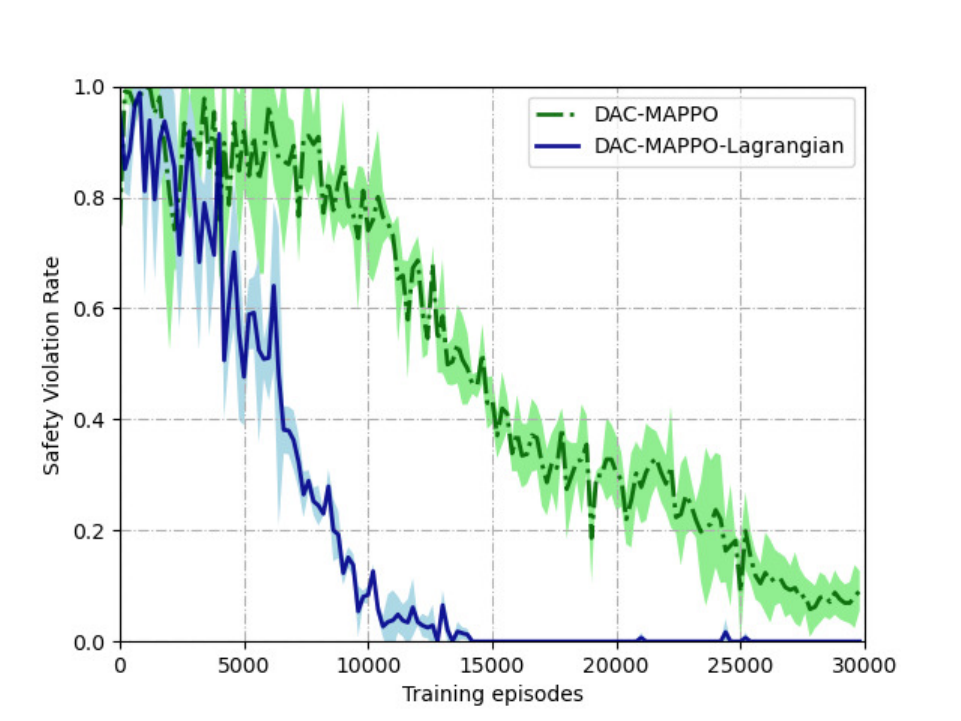}}
\caption{Training performance of DAC-MAPPO and DAC-MAPPO-Lagrangian \textcolor{black}{in Scenarios 1 and 2 (i.e., S1 and S2)}. The shaded areas denote the standard errors across three independent runs.}
\label{training-result} 
\end{figure*}

Fig.~\ref{training-result} presents the training performance of the two DRL-based algorithms, DAC-MAPPO and DAC-MAPPO-Lagrangian. During the training, the policies are periodically evaluated, with $100$ test episodes conducted every $100$ training episodes. The shaded areas indicate the standard errors across the three independent runs. In Fig.~\ref{training-returnS1} and Fig.~\ref{training-returnS2}, the X-axis shows the number of training episodes, and the Y-axis represents the average operational return across the $100$ test episodes at each evaluation point. Fig.~\ref{training-rateS1} and Fig.~\ref{training-rateS2} illustrates the evolution of the safety violation rate during training. 

In Scenario 1, DAC-MAPPO-Lagrangian demonstrates faster and more stable convergence than DAC-MAPPO. As shown in Fig.~\ref{training-returnS1}, DAC-MAPPO-Lagrangian stabilizes after approximately $8{,}000$ training episodes, while DAC-MAPPO requires about $17{,}500$ episodes to converge. This improvement stems from the primal–dual structure of the Lagrangian formulation, which adaptively balances operational performance and constraint satisfaction during training. In contrast, DAC-MAPPO relies on a manually tuned safety-weight parameter, making it more sensitive and slower to stabilize. Moreover, the shaded region for DAC-MAPPO-Lagrangian is narrower, indicating reduced variance and superior training robustness across independent runs.

{\color{black}In Scenario 2, which involves a larger fleet, both algorithms require longer training horizons due to the enlarged state and action spaces. However, the performance trend remains consistent: DAC-MAPPO-Lagrangian converges significantly faster (around $14{,}500$ episodes) compared to DAC-MAPPO (approximately $26{,}000$ episodes). This demonstrates that the Lagrangian-based optimization still accelerates convergence under large-scale and more complex environments.}

Meanwhile, as shown in Figs.~\ref{training-rateS1} and \ref{training-rateS2}, DAC-MAPPO-Lagrangian consistently maintains near-zero violation rates after convergence in both scenarios. In contrast, DAC-MAPPO exhibits noticeable fluctuations before convergence, {\color{black}particularly in Scenario 2, and stabilizes only at a much later stage. Overall, these results confirm that the proposed Lagrangian formulation significantly improves convergence efficiency, stability, and safety consistency across different fleet scales.}\par


\subsubsection{Charging Schedule Results}

\begin{figure*}[t]
	\centering 
	\subfigure[DAC-MAPPO-Lagrangian]{
        \includegraphics[height=11cm,width=0.27\linewidth]{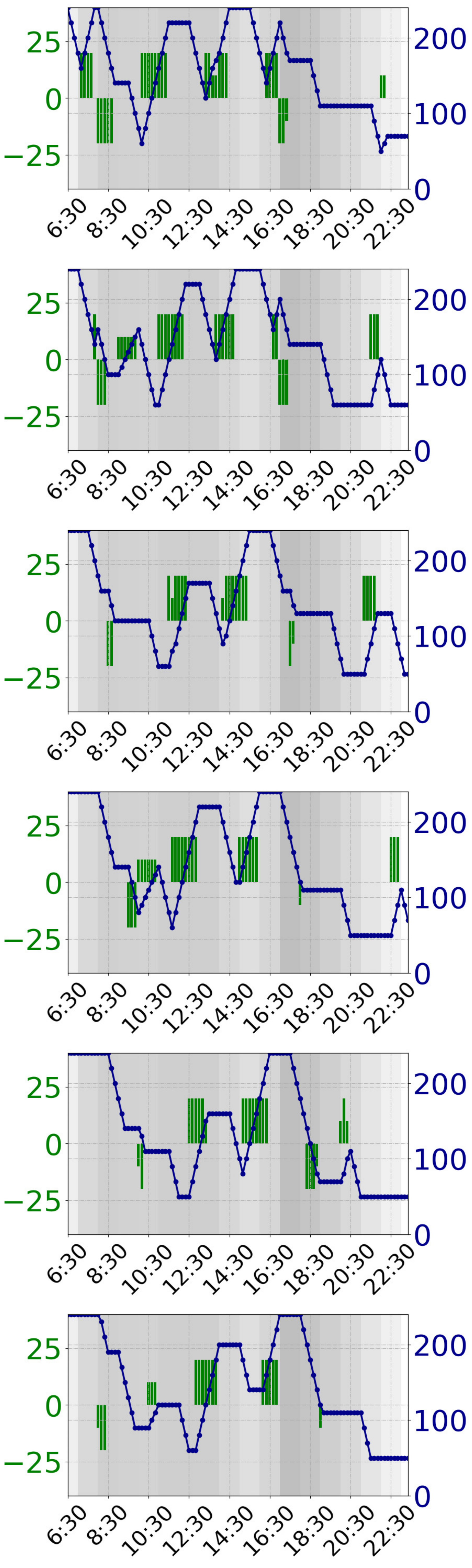}
		\label{trajectoryMAPPOLagrangian}
	}	
        \subfigure[DAC-MAPPO]{
        \includegraphics[height=11cm,width=0.27\linewidth]{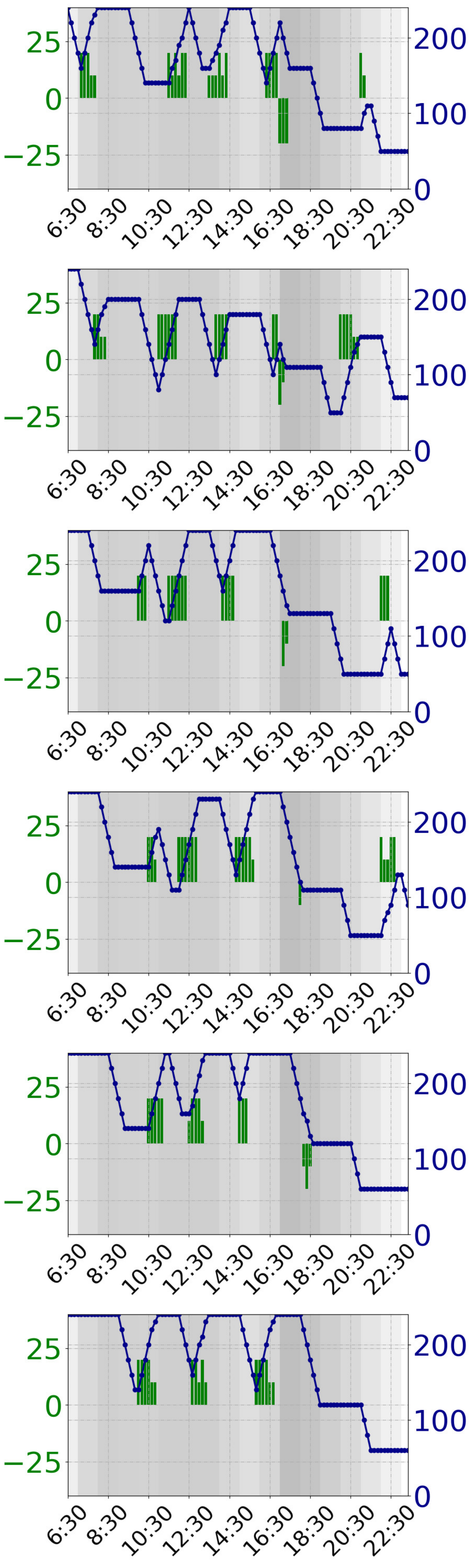}
		\label{trajectoryMAPPO}
	}
	\subfigure[MILP-S]{
        \includegraphics[height=11cm,width=0.27\linewidth]{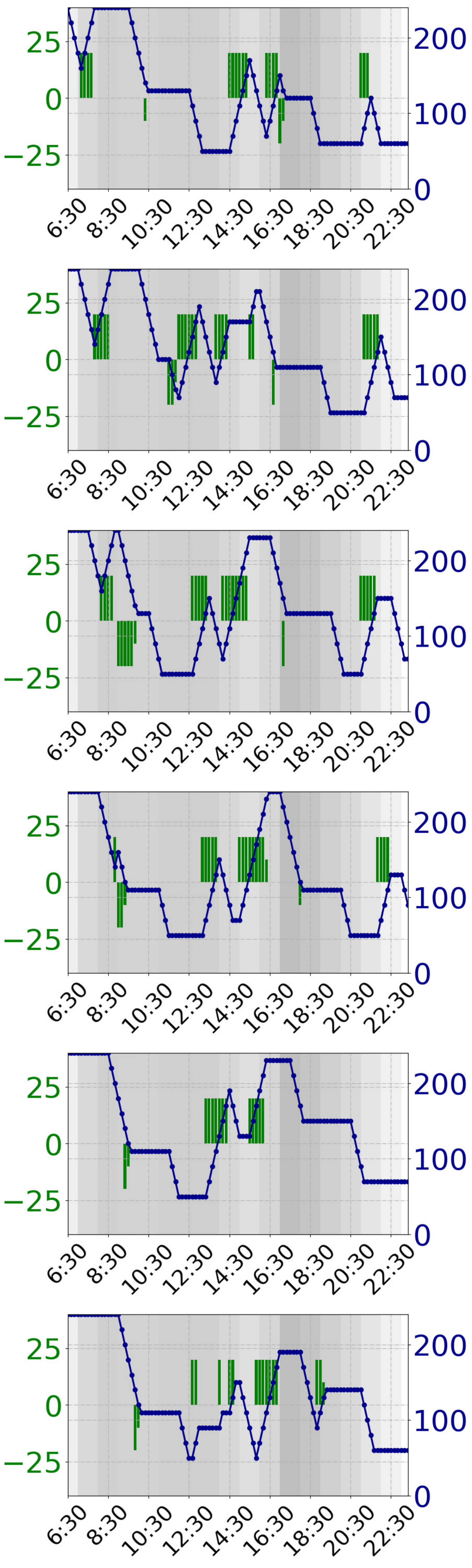}
		\label{trajectoryMILP}
	}
	\caption{The detailed charging schedules of DAC-MAPPO-Lagrangian, DAC-MAPPO, and MILP-S.}
    \label{trajectory}
\end{figure*}

To gain insights into the behavior of different charging scheduling policies, we analyze a representative episode selected from the test set, which spans from the first time step to the end of the scheduling horizon. {\color{black}Due to space limitations and the difficulty of visually presenting all trajectories for the larger fleet, only the results of Scenario~1 are illustrated here.} The corresponding PV power profile for this episode is shown in Fig. \ref{datasetPV}. As a result, Fig. \ref{trajectory} presents the charging schedules of all six EBs under three different algorithms: DAC-MAPPO-Lagrangian, DAC-MAPPO, and MILP-S. In each subfigure, the curves represent the battery level trajectories of the EBs over time, while the bars indicate the charging power decisions at each time step. Additionally, the grayscale background highlights electricity price variations, with darker shades indicating higher prices.

Firstly, DAC-MAPPO-Lagrangian demonstrates a clear ability to optimize charging strategies for maximum profitability. As shown in Fig.~\ref{trajectoryMAPPOLagrangian}, the EBs predominantly charge during two key periods: (1) when PV output is high (e.g., 11:00–14:00), and (2) when electricity prices are relatively low (e.g., 15:00–17:00). Moreover, the activities of selling electricity back to the grid are also concentrated in two intervals. The first is in the morning between 8:00 and 10:00. Although this is not the peak price period, EBs can replenish energy during the midday high-PV window, thus avoiding the risk of battery depletion. The second occurs during the actual peak price period from 17:00 to 19:00. However, the amount of selling during this interval is relatively limited, which reflects a learned trade-off: since PV generation is unavailable at night, and electricity selling involves a discount, the agent conserves some energy for later use during nighttime operations (i.e., after 19:00). In contrast, as illustrated in Fig.~\ref{trajectoryMAPPO}, DAC-MAPPO also capitalizes on the PV output to reduce energy costs, but its strategy is more conservative. Notably, none of the EBs sell electricity during the morning period. Examining the SoC trajectories reveals that this strategy maintains relatively high battery levels before 18:00, further confirming the agent's cautious behavior in the absence of explicit safety modeling.

Finally, we compare the charging schedules between MILP-S and DAC-MAPPO-Lagrangian. As shown in Fig.~\ref{trajectoryMILP}, the charging strategy of MILP-S does not align well with the real-time electricity price trends. For example, due to an underestimation of electricity prices, both EB2 and EB3 perform intensive charging between 8:00 and 9:00, a period when PV output is insufficient. As a result, electricity must be drawn from the grid, leading to increased operational costs. Furthermore, MILP-S also underestimates PV output power, and thus the EBs miss the opportunity to charge during the 9:00–10:00 window, when PV generation is available and electricity prices are moderate. This suboptimal behavior can be attributed to MILP-S's reliance on forecasted electricity prices and PV output, which may significantly deviate from actual real-time conditions. In contrast, DAC-MAPPO-Lagrangian benefits from the adaptability of DRL, enabling it to respond dynamically to uncertain environments and make more robust and cost-effective decisions in real time.

\section{Conclusion}
In this paper, we have employed safe RL and HDRL techniques to optimize charging schedules for EB fleets under realistic operating conditions with multi-source uncertainties. By incorporating dynamic PV generation, electricity pricing, and variable travel times, our system model has better reflected the complexities of sustainable public transportation systems. To handle these challenges, we have reformulated the EBCSP as a CMDP with options, and further decomposed it into two augmented CMDPs within the DAC architecture. Specifically, the high-level charger allocation actions persist over variable time periods, while the low-level charging power actions are selected at each time step. To solve these CMDPs, we have presented the DAC-MAPPO-Lagrangian algorithm, which integrates Lagrangian relaxation into both centralized PPO at the high level and decentralized MAPPO at the low level. This allows for automatic balancing between cost minimization and safety satisfaction without manual penalty tuning.\par 
Experimental results on real-world datasets have validated the superiority of the proposed method over existing baselines in terms of operational cost, safety compliance, and convergence stability. Notably, our approach achieves near-zero safety violations across test episodes, demonstrating its robustness in maintaining battery reliability and avoiding energy depletion under uncertain environments. This safety guarantee is crucial for real-world deployment, where breakdowns due to insufficient charge must be strictly avoided.\par
As future work, we aim to explore adaptive constraint thresholds and online safety monitoring mechanisms to further enhance the practicality of safe RL in large-scale, uncertainty-aware transportation systems. \par

\begin{appendices}
		\section{Proof for Proposition \ref{Pro1}}
		\begin{proof}
	From the definition of the high-level reward value function in \eqref{value_cost_high}, we have:
\begin{equation}
V_{\pi_{\mathrm{H}}}^{\mathrm{R}}(s) = \underset{\pi_\mathrm{H},\pi_\mathrm{L}}{\mathbb{E}} \left[ \sum_{t'=t}^{T-1} r_{\rm }^{\mathrm{opr}}(S_{t'}, A_{t'}) \,\middle|\, S_t = s \right]. \label{eq:high_reward_def}
\end{equation}
Then, we can marginalize over $\omega$:
\begin{align}
V_{\pi_{\mathrm{H}}}^{\mathrm{R}}(s) 
&= \sum_{\omega \in \varOmega_t} \pi_{\mathrm{H}}(\omega \mid s) \nonumber \\
&\underset{\pi_\mathrm{L}}{\mathbb{E}} \left[ \sum_{t'=t}^{T-1} r_{\rm L}^{\mathrm{opr}}(S_{t'}, \omega_{t'}, p_{t'}) \,\middle|\, S_t = s, \omega_t = \omega \right] \nonumber \\
&= \sum_{\omega \in \varOmega_t} \pi_{\mathrm{H}}(\omega \mid s) \cdot V_{\pi_{\mathrm{L}}}^{\mathrm{R}}(s, \omega),
\end{align}
which means that \eqref{value_R_highlow} holds. 

Note that the high-level safety cost is the same as the low-level safety cost, i.e., $c_{\rm H}^{\rm safe}(S_t)=c_{\rm L}^{\rm safe}(S_t)=c_{t}^{\rm safe}$, as given in Definition 1 and Definition 2. Therefore, we have:
\begin{equation}
V_{\pi_{\mathrm{H}}}^{\mathrm{C}}(s) = \underset{\pi_\mathrm{L}}{\mathbb{E}} \left[ \sum_{t'=t}^{T-1} c_{t'}^{\rm safe} \right] = V_{\pi_{\mathrm{L}}}^{\mathrm{C}}(s),
\end{equation} 
which means that \eqref{value_C_highlow} holds.
		\end{proof}
{\color{black}
        \section{Proof for Theorem \ref{dualgap}}
		\begin{proof}
Because $r_{\rm H}^{\rm opr}(s,\omega)$ and $-c_{\rm H}^{\rm safe}(s)$ are bounded, the assumptions of Theorem 1 in \cite{paternain2019constrained} are satisfied. Therefore strong duality holds for \eqref{cobjectivehigh}, i.e., $\mathbf{P}_{\rm H}^*=\mathbf{D}_{\rm H}^*$.


        		
Moreover, since the policy parameterization in \eqref{cobjectivehigh} using DNN is an $\epsilon$-universal approximator (as required by the assumptions of Theorem 2 in \cite{paternain2019constrained}), then the parameterization error is bounded, and Theorem 2 in \cite{paternain2019constrained} applies. Consequently, the parametrized duality gap bound \eqref{parametergap} holds.
\end{proof}
}
\end{appendices}

\bibliography{author}
\bibliographystyle{IEEEtran}

\end{document}